\documentclass{article}
\usepackage{xcolor} %

\usepackage[final]{corl_2026} %

\usepackage{amsmath}    %
\usepackage{amssymb} 
\usepackage[ruled,linesnumbered]{algorithm2e}
\usepackage{booktabs}
\usepackage{longtable}
\usepackage{multirow}
\usepackage{graphicx}
\usepackage{makecell}
\usepackage{caption}
\usepackage{float}
\usepackage{wrapfig}
\hypersetup{allcolors=black}

\title{EmbodiSteer: Steering Embodiment-Agnostic Visuomotor Policies with Joint-Space Guidance for Zero-Shot Cross-Embodiment Deployment}

\author{
  {\normalsize
  Shihefeng Wang$^{*,1,2,3}$\hspace{0.8em}
  Kangchen Lv$^{*,1,2,3}$\hspace{0.8em}
  Mingrui Yu$^{\dagger,1,2,3}$\hspace{0.8em}
  Xiang Li$^{\dagger,1,2,3}$} \\
  {\normalfont\small $^1$Tsinghua University,\hspace{0.8em} $^2$Beijing Key Laboratory of Embodied Intelligence Systems,\hspace{0.8em} $^3$BNRist} \\
      {\normalfont\small $^*$Equal contribution. \quad $^\dagger$Corresponding author.}
}

\begin{document}
\maketitle
\vspace{-2.5em}
\noindent
\begin{minipage}{\linewidth}
\centering
\includegraphics[width=\linewidth]{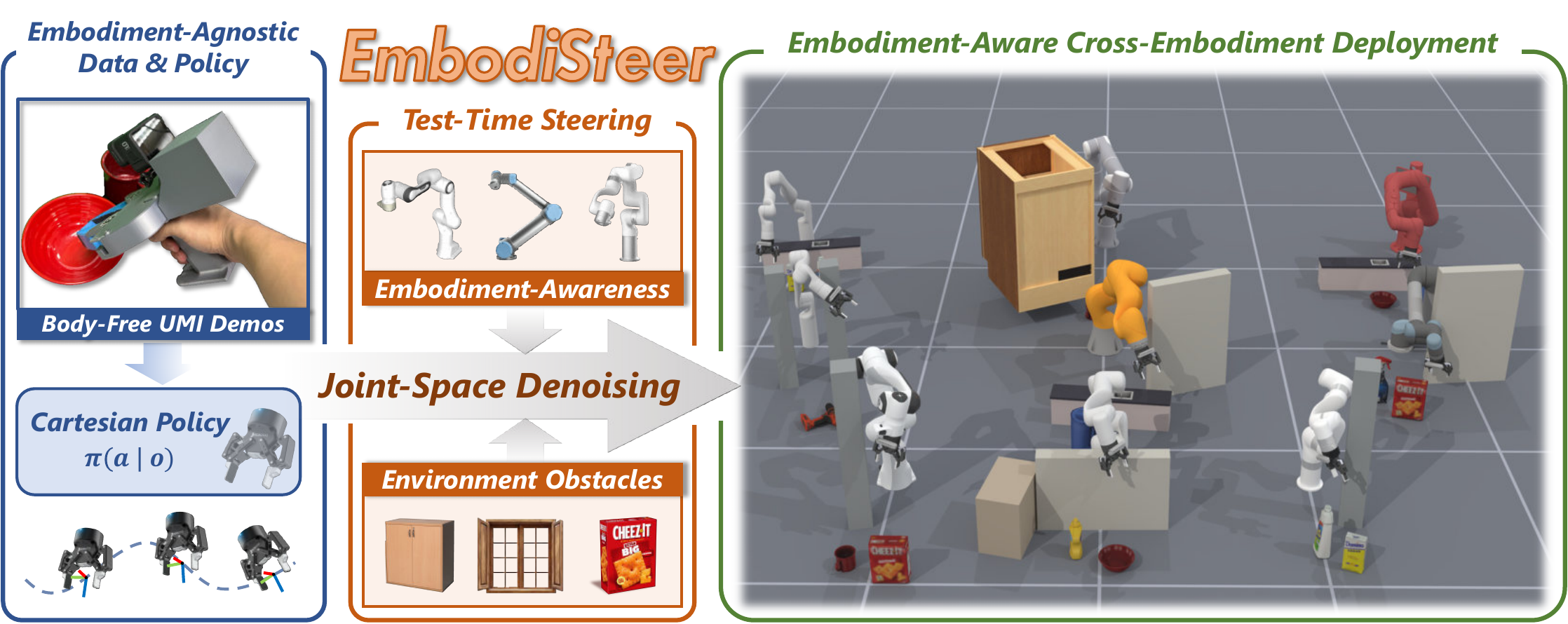}
\captionsetup{justification=justified,singlelinecheck=false}
\captionof{figure}{We present \textbf{\textsc{EmbodiSteer}}, an inference-time steering framework for embodiment-aware deployment of embodiment-agnostic visuomotor policies. Given a trained Cartesian policy (left), EmbodiSteer lifts the 
sampling process into the target robot's joint space and incorporates robot embodiment
and obstacle guidance during denoising (middle), enabling zero-shot whole-body
collision-aware execution across diverse robots without finetuning (right).}
\label{fig:top_fig}
\end{minipage}

\begin{abstract}
Scalable robot imitation learning relies on large-scale heterogeneous data from diverse robots or body-free data, making Cartesian end-effector actions a key interface for embodiment-agnostic policy learning. However, end-effector-only abstraction leaves Cartesian policies unaware of the deployed robot body, making them brittle under robot-specific constraints such as whole-body collision avoidance. To overcome this limitation, we present \textbf{\textsc{EmbodiSteer}}, a training-free framework that steers embodiment-agnostic visuomotor policies toward zero-shot, embodiment-aware deployment. \textsc{EmbodiSteer} keeps policy learning in Cartesian space while efficiently lifting inference-time diffusion sampling into the target robot’s joint space via forward kinematics and Jacobian-based updates. 
With whole-body collision-aware guidance over joint trajectories after each denoising step, the arm can be steered away from collisions while preserving learned end-effector behavior. 
Across 9 simulated robots, \textsc{EmbodiSteer} reduces collision rate by 46.1 percentage points and increases task success by 28.5 percentage points over Cartesian-only execution; on two physical robots in highly constrained scenarios, the respective improvements are 90.0 and 36.7 percentage points.
Our project page is at \url{https://frankwang67.github.io/EmbodiSteer-Page}.

\end{abstract}


\section{Introduction}

Recent progress in robot imitation learning has been driven by large-scale behavior cloning methods, from diffusion policies~\cite{chi2025diffusion,ze20243d} to vision-language-action models~\cite{zitkovich2023rt,kim2024openvla,zheng2025x,liu2025rdt}. 
One of the key factors behind this scalability is the use of Cartesian end-effector poses as a unified action representation~\cite{wang2026rethinking}, whose dimensionality and semantics are independent of a robot's degrees of freedom, joint ordering, or link geometry. 
Consequently, demonstrations from heterogeneous robots, simulators, and body-free data collection systems such as UMI~\cite{chi2024universal} can be represented in a shared action space, providing an embodiment-agnostic foundation for cross-embodiment policy learning.

However, the embodiment-agnostic property of Cartesian policies comes with a deployment-time limitation of body unawareness, which becomes problematic when execution must satisfy embodiment-specific constraints. For example, the entire arm must avoid collisions in cluttered scenes, and predicted end-effector motions may become infeasible near workspace boundaries for a specific robot. Training directly in joint space could introduce embodiment awareness, but joint actions are morphology-specific and unavailable in UMI-style body-free demonstrations, making zero-shot deployment to diverse embodiments difficult without robot-specific finetuning data.

These observations motivate us to revisit the central challenge of \textbf{cross-embodiment deployment of embodiment-agnostic policies}.
Existing research on cross-embodiment policy generalization largely treats embodiment differences as shifts in action and state distributions, addressing them through embodiment-specific action decoders~\cite{bauer2025latent,bu2025univla,wang2024scaling}, reusable low-level action primitives or skills~\cite{xu2023xskill,zha2026lap,zheng2025universal}, and embodiment-conditioned policy representations~\cite{zheng2025x,patel2025get,gupta2022metamorph}. 
However, these works leave execution-time constraints of the target robot body largely unaddressed, so end-effector motion may transfer across robots while whole-body collisions still cause failures.
In this work, we focus on the complementary problem of feasibly and safely executing an embodiment-agnostic policy on different robot arms at inference time. We argue that \textit{true cross-embodiment deployment is more than reproducing end-effector behavior across robots}, since the policy output must be realized as embodiment-aware joint-space motion that respects the target robot's whole-body constraints.

To address this problem, we propose \textbf{\textsc{EmbodiSteer}}, a training-free inference framework for embodiment-aware deployment of learned Cartesian policies. 
The policy is trained only with the end-effector actions, thus preserving its compatibility with embodiment-agnostic demonstrations. At inference, \textsc{EmbodiSteer} lifts Cartesian diffusion sampling process into the target robot's joint space, where embodiment-specific constraints can steer the sampled motion toward collision-free, robot-feasible execution. Concretely, \textsc{EmbodiSteer} maintains a target-robot joint trajectory as the sampling state, but queries the frozen denoiser in its original Cartesian action space via forward kinematics. 
Each denoised Cartesian target is tracked in joint space with damped 
Jacobian updates, while CBF-inspired whole-body guidance applies joint corrections that move the robot body away from obstacles without largely
disrupting the policy-predicted end-effector motion.

We validate \textsc{EmbodiSteer} by directly deploying trained Cartesian policies
across robot embodiments in whole-body collision-aware manipulation scenarios,
covering 9 simulated and 2 real robot embodiments.
Compared with Cartesian-only execution, \textsc{EmbodiSteer} reduces collision rate and increases task success by 46.1 and 28.5 percentage points, respectively, in simulation, while consistently outperforming post-processing variants and cost-gradient-based steering baselines. On two physical robots, the respective improvements are 90.0 and 36.7 percentage points in highly constrained manipulation tasks, without robot-specific finetuning.

Overall, our contributions are summarized as follows:

1. We propose \textsc{EmbodiSteer}, a training-free inference framework that lifts inference-time sampling of embodiment-agnostic Cartesian policies into the target robot's joint space, enabling zero-shot cross-embodiment deployment while retaining the scalability of Cartesian policy learning.

2. We introduce CBF-inspired whole-body collision-aware guidance for sampled joint trajectories, which uses small corrective joint updates to steer the robot body away from obstacles while leveraging a Jacobian-based task-space cost to preserve the learned end-effector behavior.

3. We design systematic simulation and real-world benchmarks for cross-embodiment deployment, demonstrating that \textsc{EmbodiSteer} achieves zero-shot whole-body collision-aware execution across diverse embodiments while outperforming Cartesian-only execution and other steering baselines.

\section{Related Work}
\label{sec:related_works}

\textbf{Cross-Embodiment Policy Learning.} Large-scale cross-embodiment pretraining over diverse robot data is central to
robot generalist policies, and a common strategy is to unify heterogeneous data
in a shared action space.
Cartesian end-effector poses are widely adopted for their embodiment-agnostic semantics, and also support
body-free data collection systems~\cite{chi2024universal,xu2025dexumi,ha2024umi}.
Beyond directly mixing data from different platforms or aligning coordinate frames for training, prior works have explored several forms of embodiment alignment, such as latent action spaces with embodiment-specific decoders~\cite{bauer2025latent,bu2025univla,bjorck2025gr00t}, reusable action primitives or skill abstractions~\cite{xu2023xskill,zha2026lap,zheng2025universal}, and embodiment-conditioned policy representations~\cite{sferrazza2024body,bohlinger2024one,patel2025get,zheng2025x}.
More recently, ego-centric approaches use human hand motion as a shared intermediate space 
~\cite{luo2026being,yang2025egovla,zheng2026egoscale}.
These approaches address observation, state, and action distribution shifts across embodiments at the policy-learning level. Complementing this line of work, XMoP~\cite{rath2025xmop} learns a goal-conditioned whole-body motion planner from synthetic multi-embodiment planning demonstrations and transfers to unseen robots with the same policy weights. Our work instead introduces target-robot whole-body guidance at deployment time for an existing embodiment-agnostic Cartesian manipulation policy, without replacing or finetuning it.

\textbf{Steering Diffusion Generation with Guidance.} Diffusion-based generation in robotics can be steered at inference time without modifying pretrained models, through mechanisms such as noise biasing~\cite{wagenmaker2025steering}, post-hoc filtering~\cite{liu2025bidirectional,romer2025demonstrations}, and guided sampling. While classifier-free guidance~\cite{ho2022classifier} requires guidance signals during training~\cite{reuss2023goal}, classifier guidance~\cite{dhariwal2021diffusion} can leverage inference-time gradients without retraining, and has been widely applied to robotic diffusion generation~\cite{lee2025learning,xiao2025safe,zhang2026constrained,zhong2025dexgrasp,weng2024dexdiffuser,jia2026armaware}. 
For visuomotor policies, DynaGuide~\cite{du2025dynaguidesteeringdiffusionpolices} uses a learned dynamics model to steer base policies toward complex objectives, and ITPS~\cite{wang2025inference} incorporates real-time user interactions through test-time guidance. JM2D~\cite{jung2025joint} combines pretrained diffusion planners with model-based constraint modules through joint importance sampling, without additional training. Most related to our setting, EADP~\cite{gupta2026umi} steers embodiment-agnostic policies for embodiment-aware deployment, but focuses on low-level controller tracking costs for aerial manipulators. We instead target collision-aware whole-body execution for robot arms by lifting Cartesian policies into joint-space sampling and applying collision-aware guidance.

\textbf{Collision-Aware Visuomotor Policies.} Safety-critical robot manipulation in cluttered environments requires explicit collision awareness.
Learning-based visuomotor policies can incorporate safety through post-processing mechanisms, such as control barrier functions~\cite{ames2019control,hu2025vlsa,brunke2025semantically}, model predictive filters~\cite{wabersich2021predictive,gros2020safe}, and risk estimators~\cite{zhai2026cofreevla,romer2025demonstrations}. 
For diffusion-based policies, obstacle-related costs can further steer the sampling process toward safer trajectories at inference time~\cite{li2025language,deng2025safebimanual,dastider2024apex}.
Whole-body collision avoidance has also been studied in learned motion planning: M$\pi$Nets~\cite{fishman2023motion} generates robot motion from depth point clouds, while MPD~\cite{carvalho2023motion} combines learned diffusion trajectory priors with collision-cost guidance.
Recent works address robot-body safety by encoding whole-body geometry with point clouds, keypoints, or signed distance fields and learning from collision-free demonstrations~\cite{lv2026kinematicsaware,lv2025spatial,chen2025samp,fishman2024avoid}. 
These methods learn whole-body motion representations or policies from dedicated planning or robot-body demonstrations.
In contrast, we use a pretrained embodiment-agnostic policy as the base and introduce whole-body collision constraints during zero-shot inference.

    \section{Preliminaries: Cartesian Diffusion Policy}
\label{sec:prelim}

Imitation learning aims to learn a visuomotor policy $\pi$ from expert demonstrations, which maps observations such as images and proprioception to actions. 
Diffusion Policy~\cite{chi2025diffusion} formulates action generation as a conditional sampling problem with diffusion models. 
During training, Gaussian noise is added to a clean action sample $a_0$ from the dataset according to $A_t=\sqrt{\bar \alpha_t} A_0+\sqrt{1-\bar \alpha_t}\,\epsilon_t$ at diffusion timestep $t$, where $\bar \alpha_t$ denotes noise schedule coefficient and $\epsilon_t \sim\mathcal{N}(0,I)$. The denoising model is trained to predict the added noise by minimizing $\mathcal{L}=\mathrm{MSELoss}(\epsilon_t,\epsilon_\theta(A_t,o,t))$. At inference time, starting from $A_T\sim\mathcal N(0,I)$, the sample is iteratively
denoised by the DDPM reverse update as
\begin{equation}\label{eq:denoise}
  A_{t-1}  = 
  \frac{1}{\sqrt{\alpha_t}}
\left(
A_t
-
\frac{1-\alpha_t}{\sqrt{1-\bar{\alpha}_t}}
{\epsilon}_\theta(A_t,o,t)
\right)
+
\sigma_t {z},
\quad
{z}\sim\mathcal{N}(0,I).
\end{equation}

We use Cartesian end-effector poses as the action space for
embodiment-agnostic policies, which predict an action chunk
$A=[a_1,\dots,a_H]$ over horizon $H$. Here, each action $a_i = [\delta p_i,\; r_i] \in \mathbb{R}^{9}$ is represented relative
to the chunk-start end-effector pose, where $\delta p_i\in\mathbb{R}^3$ and $r_i\in\mathbb{R}^6$ denote the relative translation and 6D rotation
representation~\cite{zhou2019continuity}, respectively. Notably, the gripper command is also included in the action but will be omitted for writing brevity in the following sections.

\begin{figure}[t]
    \centering
    \includegraphics[width=\linewidth]{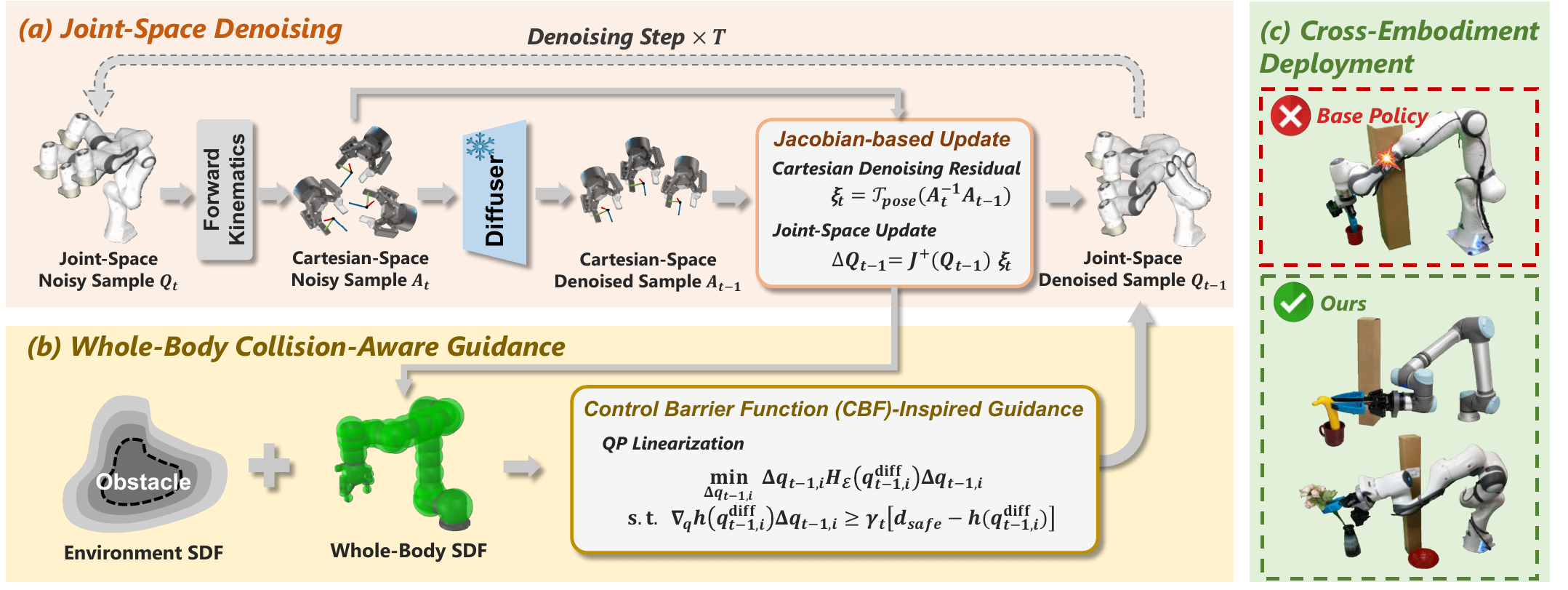}
        \caption{Overview of \textsc{EmbodiSteer}. (a) A trained Cartesian policy is lifted into the target-robot joint space for inference-time sampling. (b) CBF-inspired guidance steers sampled joint trajectories away from whole-body collisions while preserving end-effector behavior. (c) Visualization of real-world deployment across diverse robot embodiments on constrained manipulation tasks.}
    \label{fig:overview}
    \vspace{-1.0em}
\end{figure}

\section{Method}
\label{sec:method}

As shown in Fig.~\ref{fig:overview}, the Cartesian end-effector policy is trained on embodiment-agnostic demonstrations and maintained frozen during cross-embodiment inference-time steering. In this section, we first detail our problem formulation of whole-body collision-aware deployment, followed by our proposed joint-space denoising and CBF-inspired whole-body collision-aware guidance mechanism.

\subsection{Problem Formulation}
\label{sec:problem_formulation}

We consider deploying a frozen embodiment-agnostic Cartesian diffusion policy $\pi$ on a target robot embodiment $\mathcal{E}$ with obstacle information $\mathcal{O}$. The policy generates an end-effector action chunk $A=[a_1,\dots,a_H]$ in Cartesian space, but has no access to the robot's joint configuration and whole-body collision state during sampling. Therefore, the target-robot joint trajectory $Q=[q_1,\dots,q_H]$, where $q_i \in \mathbb{R}^{D}$ for robot with $D$ controllable arm joints, is required for reasoning about embodiment-specific constraints. 
The desired whole-body safety condition is
\begin{equation}
  h(q_i) = \mathrm{SDF}(\mathcal{E}(q_i),\mathcal{O}) \ge d_{\mathrm{safe}},
  \qquad
  \forall\, i=1,\dots,H,
\end{equation}
where positive signed distance indicates clearance, negative values indicate
penetration, and $d_{\mathrm{safe}}>0$ is the clearance margin.

\begin{algorithm}[t]
\caption{\textsc{EmbodiSteer}: Joint-Space Inference with Whole-Body Collision Guidance}
\label{alg:embodisteer}
\SetAlgoLined
\KwIn{Observation $o$, obstacles $\mathcal{O}$, embodiment $\mathcal{E}$}
\KwOut{Joint action chunk $Q_0$}

Sample Cartesian noise $a_{T,i}$ and initialize $Q_T=[q_{T,1},\dots,q_{T,H}]$ using ~\eqref{eq:joint_init_refactor}\;

\For{$t=T,\dots,1$}{
  \tcp{Joint Space Denoising}

  Map $Q_t$ to Cartesian action $A_t$ using $\mathrm{FK}_{\mathcal{E}}$ \tcp*{Kinematic mapping}

  $A_{t-1} \leftarrow
  \text{DiffusionStep}(\epsilon_\theta(A_t,t,o),\,t,\,A_t)$\tcp*{Frozen Cartesian policy}

  $\left[\xi_{t,i}\right] \leftarrow \mathcal{T}_{\mathrm{pose}}\left(A_t^{-1}A_{t-1}\right)$\tcp*{Cartesian denoising residual}

  Update joint space sample $Q_{t-1}^{\mathrm{diff}}$ using ~\eqref{eq:joint_space_denoising_update}\tcp*{Jacobian residual update}

  \tcp{Whole-Body Collision-Aware Guidance}

  Solve ~\eqref{eq:cbf_qp_refactor} in batch to get solution $\Delta Q_{t-1}^{\mathrm{cbf}}$\tcp*{Solve QP}
  
  Apply the solution $\Delta Q_{t-1}^{\mathrm{cbf}}$ to $Q_{t-1}^{\mathrm{diff}}$ using ~\eqref{eq:cbf_qp_update} to get $Q_{t-1}$\tcp*{Apply the guidance}
}
\Return{$Q_0$}
\end{algorithm}

\subsection{Joint-Space Denoising with Pretrained Cartesian Diffusion Policy}
\label{sec:joint_inference_refactor}
A straightforward way to lift Cartesian denoising into joint space is to solve inverse kinematics under the target robot $\mathcal{E}$ after each denoising step, i.e., $q_{t,i}=\mathrm{IK}_{\mathcal{E}}(a_{t,i})$~\cite{jia2026armaware}. %
However, repeatedly solving IK inside the denoising 
loop is costly and hinders real-time deployment. Instead, we treat the Cartesian 
denoising update as a local task-space residual and map it to joint space using 
the target robot's Jacobian $J_\mathcal{E}(q) \in \mathbb{R}^{6 \times D}$. Since adjacent denoising steps typically induce small Cartesian changes, the first-order kinematic relation provides a reliable joint-space update without IK solving at every step. Notably, the damped pseudoinverse Jacobian $J_\mathcal{E}^+(q) = J_\mathcal{E}(q)^\top \bigl(J_\mathcal{E}(q)J_\mathcal{E}(q)^\top + \lambda_{\mathrm{pinv}} I\bigr)^{-1}$ is used here to improve numerical stability and prevent large joint motions near singularities, where $\lambda_{\mathrm{pinv}}$ is the damping factor.

\textbf{Denoising Initialization.}
We initialize joint-space sampling by drawing Cartesian Gaussian noise
$a_{T,i}\in\mathbb{R}^{9}$ and locally mapping it around the chunk-start
configuration $q^{\mathrm{start}}$:
\begin{equation}
\label{eq:joint_init_refactor}
  q_{T,i} = q^\mathrm{start}
  + \alpha \cdot \mathrm{clip}\bigl(
  J_\mathcal{E}^+(q^\mathrm{start})\,\xi_{T,i},\,
  -\Delta q_\mathrm{max},\, \Delta q_\mathrm{max}
  \bigr),
  \quad
  \xi_{T,i}=\mathcal{T}_{\mathrm{pose}}(a_{T,i}).
\end{equation}
Here, $\mathcal T_{\mathrm{pose}}$ converts the Cartesian noise into a 6D pose
twist relative to the chunk-start end-effector pose, $\Delta q_\mathrm{max}$
bounds the per-joint perturbation, and $\alpha$ is a global scaling factor. This
local kinematic mapping keeps joint-space samples consistent with the Cartesian
diffusion initialization.

\paragraph{Joint-Space Denoising Update.}
At denoising step $t$, \textsc{EmbodiSteer} maintains a noisy joint action chunk $Q_t=[q_{t,1},\dots,q_{t,H}]$.  We first project $Q_t$ through forward kinematics and convert the resulting end-effector poses into the relative Cartesian action chunk $A_t$. The frozen Cartesian denoising network is then queried in its original action space, yielding the Cartesian denoising target $
  A_{t-1}
  =
  \mathrm{DiffusionStep}\bigl(\epsilon_\theta(A_t,t,o),t,A_t\bigr) $.
Each denoising update is converted into a local 
6D pose twist $\xi_{t,i}=\mathcal{T}_{\mathrm{pose}}(a_{t,i}^{-1}a_{t-1,i})$, and mapped back to the joint space with the damped Jacobian: 
\begin{equation}
\label{eq:joint_space_denoising_update}
    q_{t-1,i}^{\mathrm{diff}} =
  q_{t,i} +
  \mathrm{clip}\bigl(J_\mathcal{E}^+(q_{t,i})\xi_{t,i},\,
  -\Delta q_\mathrm{max},\, \Delta q_\mathrm{max}\bigr).
\end{equation}

\subsection{CBF-Inspired Whole-Body Collision-Aware Guidance}
\label{sec:cbf_guidance_refactor}

Maintaining the denoising state in joint space allows \textsc{EmbodiSteer} to consider whole-body constraints during sampling. A direct approach is energy-based guidance with gradients over the joint trajectory\cite{li2025language,ma2025constraint}, which penalizes unsafe configurations at each denoising step:
$q_{t-1,i}
  =
  q_{t-1,i}^{\mathrm{diff}}
  -
  \rho_t \nabla_{q}\mathcal{L}_{\mathrm{coll}}\left(q_{t-1,i}^{\mathrm{diff}}\right)$,
where $\mathcal{L}_{\mathrm{coll}}(q)$ is the cost function punishing whole-body collision, and $\rho_t$ is the guidance strength. However, because whole-body collision gradients often originate from non-end-effector links such as the elbow or forearm, pushing the robot body away from obstacles can also perturb the end-effector trajectory needed for task completion.

\paragraph{CBF-Inspired Guidance Formulation.} \textsc{EmbodiSteer} adopts a CBF-inspired collision-aware guidance to encourage whole-body safety while preserving end-effector motion. For clarity, we describe the guidance for a single
horizon step $q_{t-1,i}^{\mathrm{diff}}$, and apply the same update
independently to all steps in the horizon. We compute a guided joint
configuration $q_{t-1,i}$ by solving
\begin{equation}
\label{eq:nonlinear_guidance_refactor}
  \min_{q_{t-1,i}}\frac{1}{2}\left\|\mathrm{FK}\left(q_{t-1,i}^{\mathrm{diff}}\right)^{-1}\mathrm{FK}\left(q_{t-1,i}\right)\right\|_W^2+\frac{1}{2}\lambda_\mathrm{cbf}\left\|q_{t-1,i}-q_{t-1,i}^{\mathrm{diff}}\right\|^2, \quad
  \mathrm{s.t.}\,\,
  h(q_{t-1,i})
  \ge
  d_{\text{safe}},
\end{equation}
where $W$ is a diagonal weight matrix, and $\lambda_\mathrm{cbf}$ balances task-space preservation and joint-space regularization. The objective tracks the diffusion update in both task and joint space, while allowing the safety constraint to move the robot body away from obstacles when necessary.

\paragraph{QP Linearization for Efficient Update.} To obtain a lightweight per-step update, we linearize the forward kinematics and safety function at $q_{t-1,i}^{\mathrm{diff}}$. Let $\Delta q_{t-1,i}=q_{t-1,i}-q_{t-1,i}^{\mathrm{diff}}$.  
Substituting $\mathrm{FK}\left(q_{t-1,i}^{\mathrm{diff}}\right)^{-1}\mathrm{FK}\left(q_{t-1,i}\right)\approx J_{\mathcal{E}}\left(q_{t-1,i}^{\mathrm{diff}}\right)\Delta q_i$ and $h\left(q_{t-1,i}\right)\approx h\left(q_{t-1,i}^{\mathrm{diff}}\right)+\nabla_q h\left(q_{t-1,i}^{\mathrm{diff}}\right)^\top\Delta q_i$ into ~\eqref{eq:nonlinear_guidance_refactor} yields the single-constraint QP:
\begin{equation}
\label{eq:cbf_qp_refactor}
  \min_{\Delta q_i}\,
  \frac{1}{2}\Delta q_{t-1,i}^\top H_{\mathcal{E}}\left(q_{t-1,i}^{\mathrm{diff}}\right)\Delta q_{t-1,i}, \,\quad
  \mathrm{s.t.}\,\,
  \nabla_q h\left(q_{t-1,i}^{\mathrm{diff}}\right)^\top\Delta q_{t-1,i}
  \ge
  \gamma_t\left[d_\mathrm{safe}-h\left(q_{t-1,i}^{\mathrm{diff}}\right)\right] ,
\end{equation}
where
$H_{\mathcal{E}}\left(q_{t-1,i}^{\mathrm{diff}}\right)=J_{\mathcal{E}}\left(q_{t-1,i}^{\mathrm{diff}}\right)^\top WJ_{\mathcal{E}}\left(q_{t-1,i}^{\mathrm{diff}}\right)+\lambda_\mathrm{cbf}I$,
and $\gamma_t$ controls the strength of the constraint. Since the QP has only one linear inequality, it can be solved efficiently in
closed form or by a batched small-scale QP solver. Denoting the solution as
$\Delta q_{t-1,i}^{\mathrm{cbf}}$, the final guided sample is
\begin{equation}
\label{eq:cbf_qp_update}
  q_{t-1,i} = q_{t-1,i}^{\mathrm{diff}}
  + \mathrm{clip}(\Delta q_{t-1,i}^\mathrm{cbf}, -\Delta q_{\max}^{\mathrm{cbf}}, \Delta q_{\max}^{\mathrm{cbf}}).
\end{equation}

\begin{figure}[t]
    \centering
    \includegraphics[width=\linewidth]{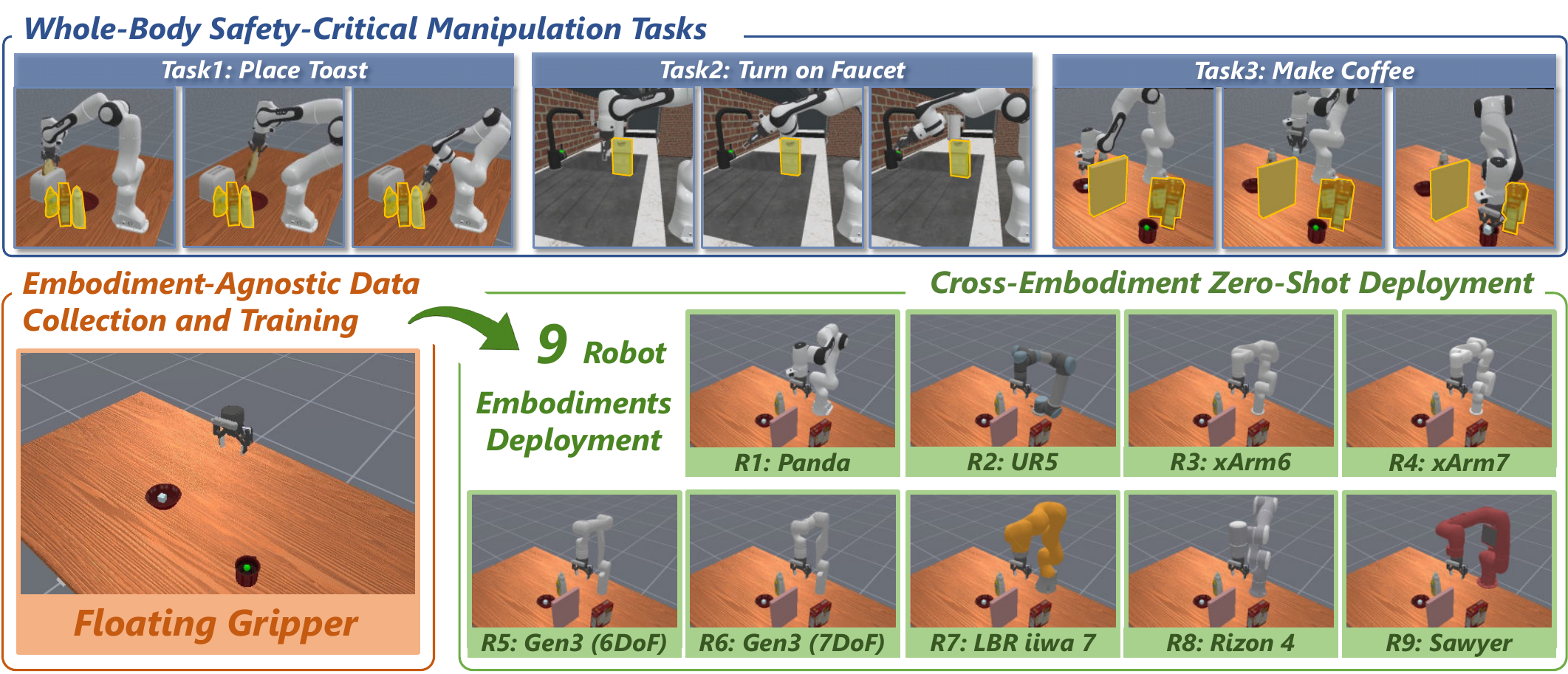}
    \caption{We evaluate \textsc{EmbodiSteer} on three manipulation tasks requiring both task completion and whole-body obstacle avoidance. Cartesian policies are trained from obstacle-free floating-gripper demonstrations and deployed zero-shot on 9 robot embodiments with test-time obstacles.}
    \label{fig:sim_exp_setup}
\end{figure}

\section{Results}

\subsection{Simulation Results}

\textbf{Experiment Setup.} We design three challenging manipulation tasks in ManiSkill~\cite{taomaniskill3}
to evaluate body-aware cross-embodiment deployment, as shown in
Fig.~\ref{fig:sim_exp_setup}. These tasks require both accurate end-effector
motion and collision avoidance between surrounding obstacles and the full robot
body. For each task, we collect 200 diverse demonstrations with a floating
gripper, recording only wrist-view RGB observations, end-effector states, and end-effector actions to keep the policy embodiment-agnostic. The trained Cartesian policy is deployed zero-shot on 9 robot arms with different morphologies, while sharing the same gripper and wrist-camera mounting. 

\textbf{Baselines.} We consider four baseline groups: 
1) \textbf{Cartesian denoising without guidance} (\textit{EE}) directly executes
the Cartesian end-effector actions predicted by the learned policy;
2) \textbf{Joint-space denoising without guidance} (\textit{Joint}) lifts
Cartesian sampling into the target robot's joint space without collision
guidance, isolating the effect of joint-space denoising alone;
3) \textbf{Cartesian denoising with post-hoc correction} leaves Cartesian
sampling unchanged and applies whole-body collision avoidance only after
trajectory generation, where \textit{EE w/ Sampling} samples multiple actions and
selects the candidate with the largest realized clearance, and \textit{EE w/
CBF} solves IK followed by a one-shot CBF-QP correction;
4) \textbf{Joint-space denoising with guidance} adds collision-aware guidance
during every joint-space denoising step, where \textit{Joint w/ CG} uses
cost-gradient guidance and \textsc{EmbodiSteer} uses our proposed task-aware CBF-QP guidance.

\begin{table}[t]
\centering
\caption{Quantitative results on manipulation tasks with obstacles, averaged over
9 robot embodiments. TSR is task success rate (\%), RWD is maximum reward per episode,
and COR is collision occurrence rate (\%). Best results among obstacle-present
methods are shown in \textbf{bold}. Repeated-seed aggregate results are reported in Tab.~\ref{tab:repeated_seed_results}.}
\label{tab:main_results}
\small
\setlength{\tabcolsep}{0pt}
\renewcommand{\arraystretch}{1.12}

\begin{tabular}{ l@{\hspace{0.45em}} |@{\hspace{0.35em}}l@{\hspace{0.45em}}|
w{c}{2.35em} w{c}{2.35em} w{c}{2.35em}
w{c}{0.85em}
w{c}{2.35em} w{c}{2.35em} w{c}{2.35em}
w{c}{0.85em}
w{c}{2.35em} w{c}{2.35em} w{c}{2.35em}
w{c}{0.3em}|w{c}{0.3em}
w{c}{2.35em} w{c}{2.35em} w{c}{2.35em} }
\toprule
\multirow{2}{*}{\textbf{Obstacles}} & \multirow{2}{*}{\textbf{Method}}
& \multicolumn{3}{c}{\textbf{PlaceToast}}
& & \multicolumn{3}{c}{\textbf{TurnOnFaucet}}
& & \multicolumn{3}{c}{\textbf{MakeCoffee}}
& & & \multicolumn{3}{c}{\textbf{Average}} \\
\cmidrule(l{0.2em}r{0.2em}){3-5}
\cmidrule(l{0.2em}r{0.2em}){7-9}
\cmidrule(l{0.2em}r{0.2em}){11-13}
\cmidrule(l{0.2em}r{0.2em}){16-18}
& & \scriptsize TSR$\uparrow$ & \scriptsize RWD$\uparrow$ & \scriptsize COR$\downarrow$
& & \scriptsize TSR$\uparrow$ & \scriptsize RWD$\uparrow$ & \scriptsize COR$\downarrow$
& & \scriptsize TSR$\uparrow$ & \scriptsize RWD$\uparrow$ & \scriptsize COR$\downarrow$
& & & \scriptsize TSR$\uparrow$ & \scriptsize RWD$\uparrow$ & \scriptsize COR$\downarrow$ \\
\midrule
\multirow{2}{*}{w/o Obs.} & EE
& 96.8 & .980 & --
& & 83.6 & .937 & --
& & 89.7 & .932 & --
& & & 90.0 & .950 & --  \\
& Joint
& 94.4 & .965 & --
& & 87.9 & .957 & --
& & 89.7 & .929 & -- 
& & & 90.7 & .950 & --\\
\midrule
\multirow{5}{*}{w/ Obs.} & EE
& 43.4 & .634 & 52.6
& & 41.6 & .752 & 61.8
& & 22.2 & .456 & 58.3
& & & 35.7 & .614 & 57.6 \\
& EE w/ Sampling
& 47.1 & .661 & 51.3
& & 43.9 & .767 & 53.2
& & 24.1 & .478 & 60.0
& & & 38.4 & .635 & 54.8 \\
& EE w/ CBF
& 65.8 & .764 & 15.3
& & 56.3 & .797 & 29.9
& & 48.9 & .606 & 8.6
& & & 57.0 & .722 & 17.9 \\
\cmidrule(l{0.2em}){2-18}
& Joint w/ CG
& 57.7 & .687 & \textbf{0.2}
& & 10.3 & .403 & 59.6
& & 22.4 & .424 & 40.6
& & & 30.1 & .505 & 33.5\\
& \textsc{EmbodiSteer}%
& \textbf{74.8} & \textbf{.829} & 4.6
& & \textbf{60.6} & \textbf{.833} & \textbf{27.1}
& & \textbf{57.2} & \textbf{.670} & \textbf{2.8}
& & & \textbf{64.2} & \textbf{.777} & \textbf{11.5} \\
\bottomrule
\end{tabular}
\vspace{-0.5em}
\end{table}

\begin{figure}[t]
    \centering
    \includegraphics[width=0.95\linewidth]{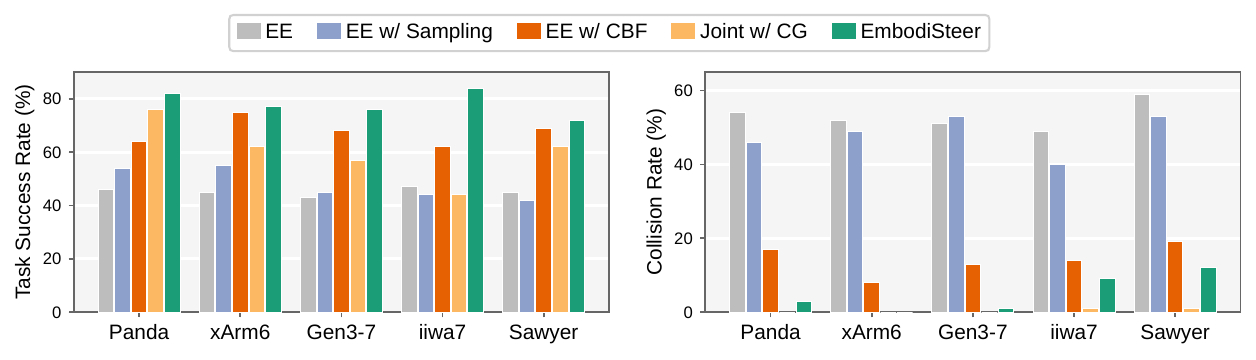}
    \caption{Per-embodiment comparison on the \textsc{PlaceToast} task, where the task success rate (left) and collision rate (right) across five representative
robot embodiments are reported.}
    \label{fig:two_figs}
    \vspace{-1.0em}
\end{figure}

\textbf{Results and Analysis.} 
Tab.~\ref{tab:main_results} reports quantitative results averaged over 9 robot embodiments. In obstacle-free settings, the Cartesian policy transfers directly across diverse arms with over 90\% average task success.
Joint-space denoising preserves task success and even yields a slight improvement, showing that sampling in joint space maintains the learned end-effector behavior. With obstacles, however, Cartesian execution drops to 35.7\% success and 0.614 reward, with 57.6\% collision rate, revealing the limitation of end-effector-only deployment under whole-arm collision constraints. Repeating the obstacle-present evaluations with two additional random seeds yields the three-seed aggregate mean $\pm$ standard deviation in Tab.~\ref{tab:repeated_seed_results}.

Among obstacle-aware methods, \textsc{EmbodiSteer} achieves the best overall balance between task success and whole-body collision avoidance.
Fig.~\ref{fig:two_figs} further compares success rate and collision rate across different embodiments, showing consistent gains over baselines. \textit{EE w/ Sampling} only modestly improves success by 2.7 percentage points and reduces collision failures by 2.8 percentage points, while incurring much higher cost from repeated sampling and collision checking.
\textit{EE w/ CBF} provides a stronger post-hoc correction, improving success by over 20 percentage points, but applies safety only after denoising and cannot shape the generated trajectory. In contrast, \textsc{EmbodiSteer} injects guidance at every joint-space denoising step, producing safer and more feasible trajectories.

  \begin{wrapfigure}{R}{0.39\linewidth}

    \vspace{-1.0em}
    \centering
    \includegraphics[width=\linewidth]{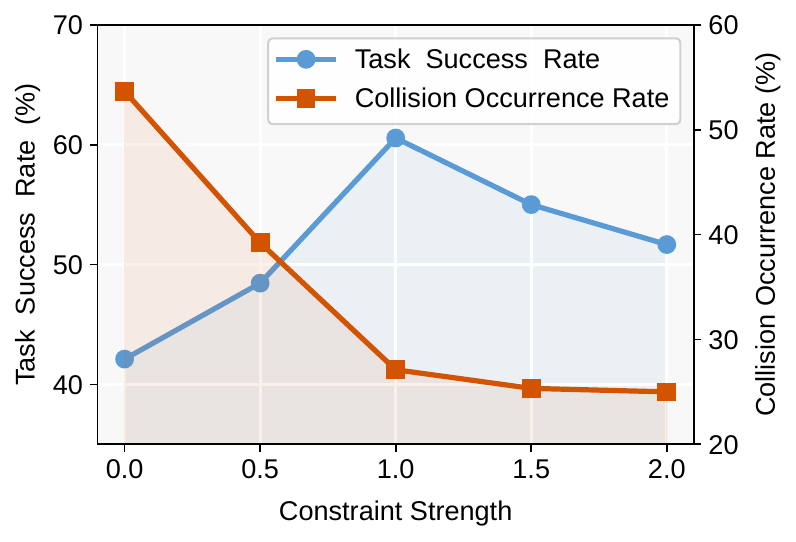}
    \caption{Ablation for the constraint strength on \textsc{TurnOnFaucet}.}
    \label{fig:guidance_scale}
    \vspace{-1.0em}

  \end{wrapfigure}

\textit{Joint w/ CG} exhibits large task-dependent variation. It performs
competitively on \textsc{PlaceToast}, where coarse end-effector motion is often
sufficient, but struggles on \textsc{TurnOnFaucet} and \textsc{MakeCoffee}, which
require precise interactions. Although collision-cost gradients can push the arm
away from obstacles, they can also significantly distort the end-effector
trajectory.
This highlights the benefit of  \textsc{EmbodiSteer}'s
CBF-QP guidance, which moves the robot body
while minimizing deviation from the policy-predicted Cartesian motion.
Finally, the constraint strength (the scale of $\gamma$) ablation shows the trade-off between safety and task
preservation. With $\gamma=0$, \textsc{EmbodiSteer} reduces to joint-space denoising without
collision guidance. Increasing $\gamma$ initially reduces collisions and
improves success, but overly large guidance eventually lowers task success by
driving samples away from the learned policy distribution.

\begin{table}[t]
\centering

\caption{Simulation results with obstacles, averaged over three tasks and
nine robot embodiments. Entries report aggregate mean $\pm$ standard deviation
across three evaluation seeds.}
\label{tab:repeated_seed_results}
\small
\setlength{\tabcolsep}{8pt}
\renewcommand{\arraystretch}{1.08}
\begin{tabular}{lccc}
\toprule
\textbf{Method} & \textbf{TSR (\%) $\uparrow$} & \textbf{RWD $\uparrow$} & \textbf{COR (\%) $\downarrow$} \\
\midrule
EE & $36.0\pm0.2$ & $.614\pm.004$ & $56.8\pm0.9$ \\
EE w/ Sampling & $37.7\pm0.5$ & $.621\pm.012$ & $53.1\pm1.3$ \\
EE w/ CBF & $57.2\pm0.6$ & $.726\pm.005$ & $18.3\pm0.3$ \\
Joint w/ CG & $30.3\pm0.5$ & $.503\pm.003$ & $33.5\pm0.6$ \\
\textsc{EmbodiSteer} & $\mathbf{63.3\pm1.4}$ & $\mathbf{.771\pm.011}$ & $\mathbf{11.7\pm0.8}$ \\
\bottomrule
\end{tabular}
\end{table}

\textbf{Alternative Constraint-Integration Baseline.}
To evaluate another training-free strategy for combining the frozen Cartesian
policy with whole-body constraints, we implement a JM2D-style
baseline~\cite{jung2025joint} that jointly samples trajectories and collision
corrections through repeated candidate denoising. Both methods use the same
frozen Cartesian policy. On \textsc{PlaceToast} with Panda, under identical
RTX 4090, batch-1, CUDA-synchronized conditions, the adaptation requires
8.65 s per chunk with 16 proposals and 16 denoising steps, compared with
0.254 s for \textsc{EmbodiSteer}. Even reducing JM2D to four steps requires
3.27 s per chunk. This computational cost limits the practicality of the
adaptation for closed-loop replanning in our setting, so we report its
computational feasibility rather than a full task-performance comparison.\par

\begin{figure}[t]
\centering
\includegraphics[width=0.83\linewidth]{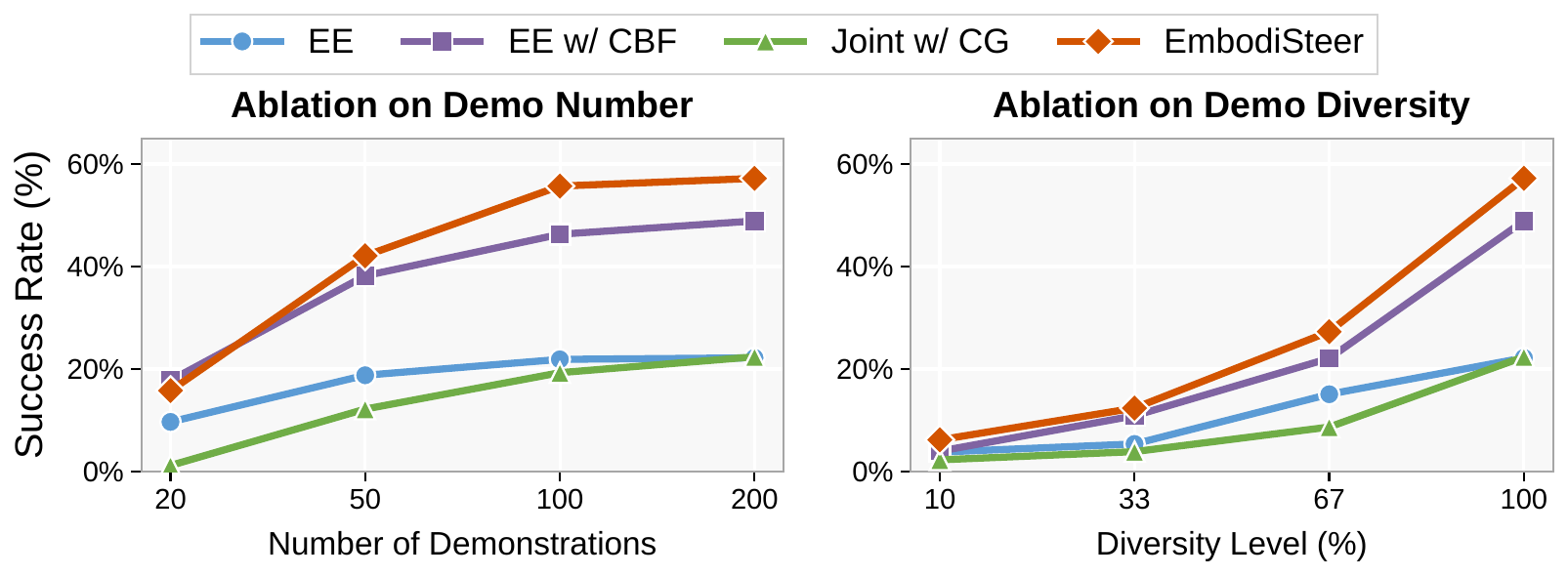}

\caption{Demonstration coverage on \textsc{MakeCoffee}: varying the
number of demonstrations (left) and motion-planning waypoint randomization
relative to the original data-generation setting (right).}
\label{fig:demo_coverage}
\end{figure}

\textbf{Demonstration Coverage.}
Collision guidance perturbs the motions proposed by the Cartesian policy,
so task performance under steering also depends on the behaviors covered by
its demonstrations. We examine this dependence on \textsc{MakeCoffee} by
retraining the Cartesian policy with 20, 50, 100, or 200 demonstrations.
Separately, we vary trajectory diversity by scaling the waypoint randomization
used during motion-planning data generation to 10\%, 33\%, 67\%, or 100\%
of the original setting, keeping 200 demonstrations at each diversity level
(Fig.~\ref{fig:demo_coverage}).
For each training setting, we compare unguided execution with the same
guidance strategies used in the main benchmark. Increasing either quantity
or diversity improves task success, with the broader coverage helping the
policy complete tasks under guidance-induced motion perturbations.
Additional obstacle-related evaluations are provided in
Appendix~\ref{app:obstacle_experiments}.\par

\subsection{Real-World Experiments}

\begin{figure}[t]
    \centering
    \includegraphics[width=\linewidth]{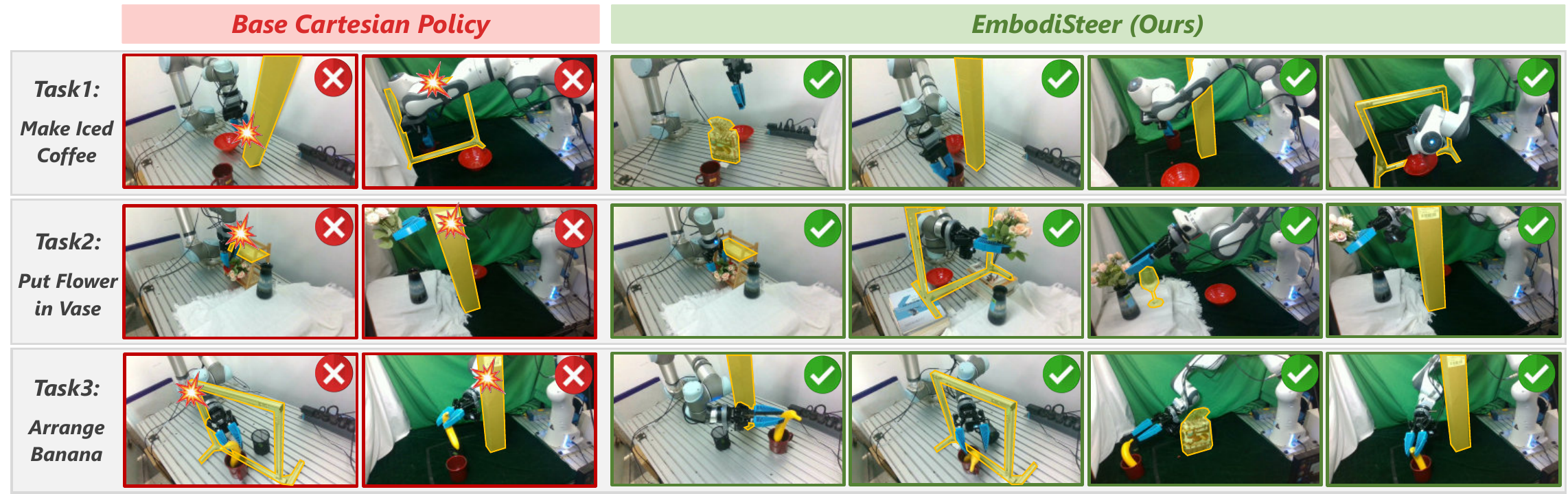}
    \caption{Real-world deployment of an arm-agnostic Cartesian policy on UR5 and Panda. Without guidance, the base policy often preserves task intent but collides with obstacles (yellow region), whereas \textsc{EmbodiSteer} steers joint trajectories to avoid arm-body collisions and complete tasks.}
    \label{fig:real_world_qual}
    \end{figure}

\textbf{Experiment Setup.}
We further evaluate \textsc{EmbodiSteer} on two physical embodiments, UR5 and Franka Panda, across
three tabletop manipulation tasks, as shown in Fig.~\ref{fig:real_world_qual}. For each task, we collect 200 demonstrations without extra obstacles, with the UMI handheld-gripper ~\cite{chi2024universal} and train a Cartesian diffusion policy without robot-specific
demonstrations or finetuning. Obstacles are introduced only at deployment time, thus directly evaluating whether \textsc{EmbodiSteer} can introduce embodiment-aware collision avoidance to an embodiment-agnostic policy.
\textsc{EmbodiSteer} also supports real-time physical deployment, achieving an inference rate of approximately 9.7 Hz on an RTX 4070 Ti SUPER.

\textbf{Results.}
For each robot and task, we conduct 10 no-obstacle trials to evaluate base task competence and 10 obstacle trials to evaluate whole-body collision avoidance. The no-obstacle setting
confirms that trained embodiment-agnostic Cartesian policies transfer to both
physical embodiments, achieving 49/60 successes. With obstacles, however, the
base policy drops to 25/60 successes and collides in 58/60 trials, showing that
end-effector-only policies lack reliable embodiment-aware execution. In contrast,
\textsc{EmbodiSteer} recovers performance to 47/60 successes while reducing collisions to 4/60.
Qualitative comparisons in Fig.~\ref{fig:real_world_qual} further illustrate how
\textsc{EmbodiSteer} improves collision-aware deployment while maintaining practical control
rates across robots and tasks.

\section{Limitations \& Conclusion }
\label{sec:conclusion}
\textbf{Limitations.} \textsc{EmbodiSteer} has several limitations. First, our guidance linearizes the whole-body constraint into a local QP and applies clipped joint corrections for stable execution; therefore, it provides an efficient feasible direction for collision avoidance but does not guarantee globally collision-free motion. Second, the guidance is activated only when the robot enters the SDF margin, making it a local correction mechanism rather than a global planner that can select the optimal collision-free motion modes far in advance. Finally, strong guidance may steer the robot into states outside the training distribution of the Cartesian policy, where task execution can fail, suggesting that broader and more diverse demonstrations remain important for robust deployment.

\textbf{Conclusions.} We propose \textsc{EmbodiSteer} for zero-shot embodiment-aware deployment of Cartesian diffusion policies by lifting inference-time sampling into the target robot's joint space and applying task-preserving whole-body collision guidance. By keeping policy learning in an embodiment-agnostic Cartesian action space while enforcing robot-specific constraints only at deployment time, the method preserves the scalability of Cartesian imitation learning and adds the body awareness needed for safe execution. Across simulation and real-world experiments, \textsc{EmbodiSteer} improves task success and substantially reduces collisions over Cartesian execution and post-hoc guidance baselines, demonstrating a practical path toward reusable manipulation policies across diverse robot embodiments. 

\clearpage
\acknowledgments{We thank Yongpeng Jiang for insightful discussions, Ting-Yu Tsui for assistance with the hardware setup, and Chendong Xin for help with the experiments. We also thank the anonymous reviewers and the Area Chair for their constructive feedback.}

\bibliography{references}  %

\clearpage
\appendix

\section{Method Details}
\label{app:method_details}

\subsection{Cartesian Action Conversion and Pose-Twist Mapping}
\label{app:pose_twist}

The main text writes the Cartesian denoising residual compactly as
$\mathcal{T}_{\mathrm{pose}}(a_{t,i}^{-1}a_{t-1,i})$. Here we make this notation
explicit for the 9D pose component of the Cartesian action. Let the chunk-start
end-effector pose be $T_0=(p_0,R_0)$ and let
$a_i=[\delta p_i,r_i]$, where $r_i\in\mathbb{R}^6$ is the
continuous 6D rotation representation. We convert $a_i$ into a
world-frame pose through
\begin{equation}
\label{eq:app_action_decode}
  \Phi(a_i;T_0)=(p_i,R_i), \qquad
  p_i=p_0+R_0\delta p_i, \qquad
  R_i=R_0\phi_{\mathrm{6D}}(r_i),
\end{equation}
where $\phi_{\mathrm{6D}}(\cdot)$ maps the 6D rotation representation to
$SO(3)$~\cite{zhou2019continuity}. The shorthand
$a_{t,i}^{-1}a_{t-1,i}$ in the main text therefore denotes the relative
transform
\begin{equation}
\label{eq:app_action_relative_transform}
  \Phi(a_{t,i};T_0)^{-1}
  \Phi(a_{t-1,i};T_0)
  =
  (\Delta p_{t,i},\Delta R_{t,i}).
\end{equation}
The pose-twist operator then returns
\begin{equation}
\label{eq:app_pose_twist}
  \xi_{t,i}
  =
  \mathcal{T}_{\mathrm{pose}}
  \left(\Phi(a_{t,i};T_0)^{-1}
  \Phi(a_{t-1,i};T_0)\right)
  =
  \begin{bmatrix}
  \Delta p_{t,i}\\
  \mathrm{Log}(\Delta R_{t,i})
  \end{bmatrix},
\end{equation}
where $\mathrm{Log}:SO(3)\rightarrow\mathbb{R}^3$ is the rotation-vector map.

The joint-space initialization follows ~\eqref{eq:joint_init_refactor}. In
experiments, $q^{\mathrm{start}}$ is the current robot joint configuration at
the beginning of the action chunk. Cartesian Gaussian noise is converted
relative to $T_0$, represented as a pose twist, and locally mapped to joint
perturbations with the damped Jacobian pseudo-inverse. During denoising, the local kinematic
approximation $\xi_{t,i}\approx J_{\mathcal E}(q_{t,i})\Delta q_{t,i}$ yields
~\eqref{eq:joint_space_denoising_update}. The per-joint clipping in
Eqs.~\eqref{eq:joint_init_refactor} and~\eqref{eq:joint_space_denoising_update}
is used to avoid large local linearization errors and unstable motions near
singularities.

\subsection{Whole-Body SDF Representation}
\label{app:sdf_details}

The whole-body safety function $h(q)$ is computed through cuRobo's collision
geometry and signed-distance interfaces. The robot body is approximated by a set
of collision spheres attached to the robot links, and we query the SDF value and
gradient of these spheres with respect to known obstacle geometry. Let
$d_j(q)$ denote the signed distance returned for the $j$-th sphere
at configuration $q$, where the convention is that positive values indicate
penetration and negative values indicate clearance. Under this convention, the
exact worst signed distance over all robot collision spheres is
\begin{equation}
\label{eq:app_exact_curobo_sdf}
  d_{\max}(q)=\max_j d_j(q).
\end{equation}
Equivalently, under the paper's convention that positive $h(q)$ indicates
clearance and negative values indicate penetration, the exact whole-body
clearance is $h_{\mathrm{exact}}(q)=-d_{\max}(q)$.

Using the exact max gives the correct worst-sphere value but can introduce
abrupt gradient switches when different collision spheres become the maximum.
For smoother guidance gradients, we use a normalized smooth
maximum over only the top-$k$ largest cuRobo signed distances. Let
$\mathcal{K}(q)$ be the indices of the $k$ largest values in
$\{d_j(q)\}_j$, and let
$m(q)=\max_{j\in\mathcal K(q)}d_j(q)$. We compute
\begin{equation}
\label{eq:app_topk_smoothmax}
  \widetilde d_{\max}(q)
  =
  m(q)
  +
  \frac{1}{\tau}
  \log\left(
  \frac{1}{k}
  \sum_{j\in\mathcal K(q)}
  \exp\left(\tau\left[d_j(q)-m(q)\right]\right)
  \right),
\end{equation}
where $\tau$ is the smoothmax temperature. The CBF safety value used by the QP is
\begin{equation}
\label{eq:app_cbf_h_from_curobo}
  h(q)=-\widetilde d_{\max}(q).
\end{equation}
Because ~\eqref{eq:app_topk_smoothmax} is a smooth approximation to the exact
max over spheres, it can introduce a small bias in the estimated worst distance.
We use it because the smoother gradient is more stable for local guidance than
the discontinuous gradient of the exact max. If the opposite SDF sign convention
is used, the same idea becomes a top-$k$ smooth minimum over clearances.

\begin{figure}[h]
\centering
\includegraphics[width=0.8\linewidth]{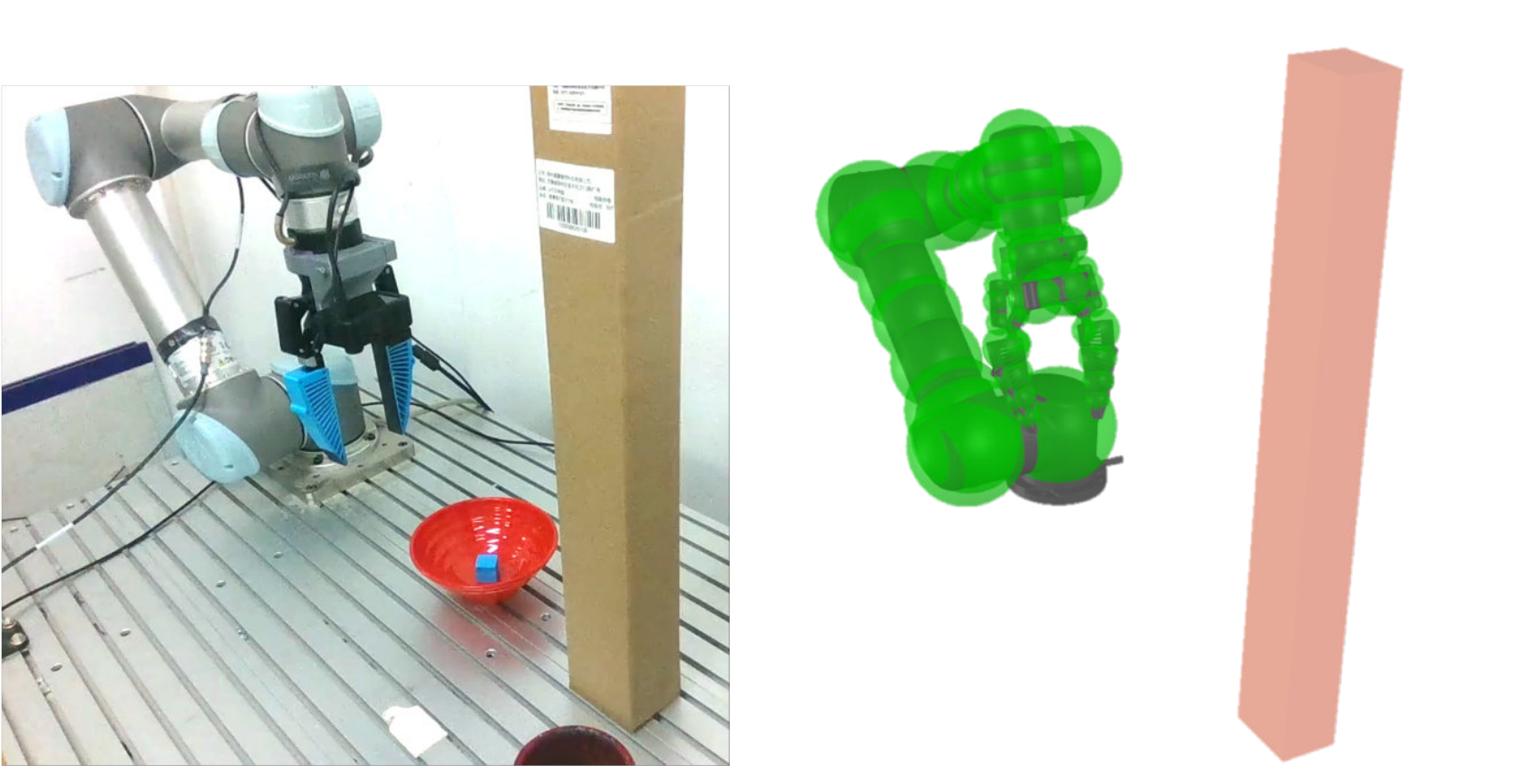}
\caption{Whole-body SDF representation. The left image shows the real scene.
The right image shows the corresponding cuRobo scene used for SDF queries,
where the robot is represented by collision spheres and the obstacle is
represented as a cuboid.}
\label{fig:app_collision_spheres}
\end{figure}

In the experiments, obstacles are modeled as known cuboids in both simulation
and real-world evaluation. We therefore assume access to obstacle pose and shape
during deployment; perception of unknown obstacles is outside the scope of this
work. Fig. ~\ref{fig:app_collision_spheres} illustrates the representation:
the physical obstacle layout is converted into known cuboids, while the robot
body is approximated by link-attached collision spheres for whole-body SDF
queries.

\subsection{CBF-Inspired QP Details}
\label{app:cbf_qp_details}

\begin{figure}[h]
\centering
\includegraphics[width=0.55\linewidth]{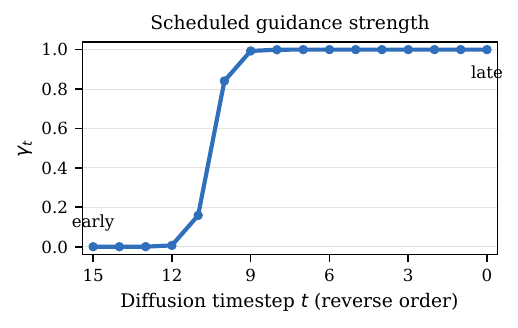}
\caption{Scheduled constraint strength $\gamma_t$ over reverse denoising steps.
The guidance is weak in early noisy steps and approaches the base scale
$\gamma$ in later steps.}
\label{fig:app_gamma_schedule}
\end{figure}

The main text derives the CBF-inspired nonlinear guidance objective and its
linearized QP form. Here we provide additional details omitted for space.
For each horizon step, ~\eqref{eq:cbf_qp_refactor} can be written as
\begin{equation}
\label{eq:app_cbf_qp_standard}
  \min_{\Delta q}\frac{1}{2}\Delta q^\top H\Delta q,
  \qquad
  \mathrm{s.t.}\quad a^\top\Delta q\ge b,
\end{equation}
where
\begin{equation}
  a=\nabla_q h(q_{t-1,i}^{\mathrm{diff}}), \qquad
  b=\gamma_t\left[d_{\mathrm{safe}}-h(q_{t-1,i}^{\mathrm{diff}})\right],
  \qquad
  H=J_{\mathcal E}^{\top}WJ_{\mathcal E}+\lambda_{\mathrm{cbf}}I.
\end{equation}
The scalar $\gamma_t$ is the scheduled constraint strength at diffusion
timestep $t$. Consistent with the reverse denoising update in the main text,
$t$ decreases from $T-1$ to $0$ during inference. Following~\cite{ma2025constraint}, we use
\begin{equation}
\label{eq:app_gamma_definition}
  \gamma_t
  =
  \gamma\cdot
  \sigma\!\left(\beta\cdot\left(c-\frac{t}{T-1}\right)\right),
  \qquad
  \sigma(z)=\frac{1}{1+\exp(-z)},
\end{equation}
where $\gamma$ is the base constraint strength. We set the slope $\beta$ to $50$ and
the transition point $c$ to $0.7$ in all experiments. This schedule gives
smaller constraint strength in early denoising steps and approaches
$\gamma$ in later steps, when samples are closer to the data
manifold and less likely to be washed out by subsequent denoising updates.
Fig.~\ref{fig:app_gamma_schedule} visualizes this schedule for the default
$T=16$ denoising steps and $\gamma=1.0$.
The QP uses the nonnegative right-hand side $\max(b,0)$, so configurations
already satisfying the safety margin receive zero correction from the constraint.

The closed-form solution follows from the KKT conditions of the single-constraint
QP. With the clipped right-hand side $\bar b=\max(b,0)$, the Lagrangian is
\begin{equation}
  \mathcal{L}(\Delta q,\mu)
  =
  \frac{1}{2}\Delta q^\top H\Delta q
  +\mu(\bar b-a^\top\Delta q),
  \qquad \mu\ge 0.
\end{equation}
Stationarity gives $H\Delta q-\mu a=0$, hence
$\Delta q=\mu H^{-1}a$. If $\bar b=0$, the unconstrained minimizer
$\Delta q=0$ is feasible. Otherwise, complementary slackness makes the
constraint active, so
$a^\top\Delta q=\mu a^\top H^{-1}a=\bar b$ and
$\mu=\bar b/(a^\top H^{-1}a)$. The resulting solution is
\begin{equation}
\label{eq:app_closed_form_qp}
  \Delta q^\star
  =
  \begin{cases}
  0, & b\leq 0,\\[0.6em]
  \dfrac{b}{a^\top H^{-1}a+\varepsilon}\,H^{-1}a, & b>0,
  \end{cases}
\end{equation}
where $\varepsilon$ is a small numerical stabilizer. The case $b\leq 0$
corresponds to a locally satisfied safety constraint, for which the
unconstrained minimizer $\Delta q=0$ is feasible. The update is then clipped as
in ~\eqref{eq:cbf_qp_update}. The QP is solved over the arm joints only; the
gripper command is not modified by the collision-avoidance correction. The task-preservation metric uses the
end-effector geometric Jacobian with separate position and rotation weights.
Because the method uses a local linearization and clipped updates, the guidance
should be interpreted as an efficient feasible correction direction rather than
a formal global safety certificate.

\section{Implementation Details}
\label{app:implementation}

\subsection{Diffusion Policy Architecture}
\label{app:policy_architecture}

We use the same Cartesian diffusion-policy architecture across all embodiments. In both
simulation and real-world experiments, RGB observations are encoded with the
pretrained CLIP ViT-B/16 backbone
(\texttt{vit\_base\_patch16\_clip\_224.openai}) loaded through timm and
finetuned during policy training. The resulting observation feature is used as a
global condition for a conditional 1D U-Net denoiser over the action chunk. The
denoiser takes a noisy Cartesian action sequence and a diffusion timestep as
input, and predicts the noise $\epsilon_\theta$ used by the reverse diffusion step. At deployment, we use a DDIM scheduler with 16 denoising steps.
The action sequence has horizon $H$ and action dimension $10$, consisting of
relative end-effector translation, a continuous 6D rotation representation, and
one gripper command. During deployment,
\textsc{EmbodiSteer} keeps this trained denoiser unchanged and only changes the
sampling variable from Cartesian actions to target-robot joint trajectories.

\subsection{Hyperparameters}
\label{app:hyperparams}

Table~\ref{tab:hyperparams} lists the hyperparameters used for training and
deployment.

\begin{table}[h]
\centering
\caption{Training and deployment hyperparameters. }
\label{tab:hyperparams}
\small
\setlength{\tabcolsep}{5pt}
\renewcommand{\arraystretch}{1.12}
\begin{tabular}{p{0.27\linewidth}p{0.18\linewidth}p{0.48\linewidth}}
\toprule
\textbf{Quantity} & \textbf{Value} & \textbf{Notes} \\
\midrule
Observation horizon & 2 & Diffusion policy input setting. \\
Action horizon $H$ & 16 & Predicted action chunk length. \\
Execution horizon & 16 in simulation, 6 in real world. & Number of actions executed before replanning. \\
Visual backbone & CLIP ViT-B/16 & \texttt{vit\_base\_patch16\_clip\_224.openai} loaded through timm, pretrained and finetuned. \\
U-Net channels & [256, 512, 1024] & Conditional 1D U-Net downsampling dimensions. \\
Diffusion step embedding & 128 & Timestep embedding dimension. \\
Diffusion training timesteps & 50 & Squared-cosine beta schedule and epsilon prediction. \\
Inference denoising steps & 16 & DDIM reverse diffusion steps at deployment. \\
Control frequency & 10 & Simulation and real-world control rate. \\
Training epochs & 120 & Used for both simulation and real-world policies. \\
Batch size & 128 in simulation, 64 in real world. & Training batch size for simulation and real-world checkpoints. \\
Optimizer & AdamW & Learning rate $3\times10^{-4}$, betas $(0.95,0.999)$, weight decay $10^{-6}$. \\
Learning-rate schedule & Cosine & 2000 warmup steps. \\
EMA & 0.9999 max decay & EMA enabled during training. \\
Image augmentation & Crop + color jitter & Random crop ratio 0.95; brightness 0.3, contrast 0.4, saturation 0.5, hue 0.08. \\
Noise initialization scale $\alpha$ & 0.1 & Scale for Jacobian-projected noise. \\
Jacobian damping $\lambda_{\mathrm{pinv}}$ & 0.001 & Used in $J_{\mathcal E}^+$. \\
Joint update clip $\Delta q_{\max}$ & 0.5 & Used in joint denoising update. \\
Safety margin $d_{\mathrm{safe}}$ & 0.05$\sim$0.10 & Task-dependent; set to 0.05, 0.07, or 0.10 by obstacle layout. \\
SDF aggregation top-$k$ & 4 & Number of critical collision spheres. \\
Smoothmax temperature $\tau$ & 20.0 & Temperature in ~\eqref{eq:app_topk_smoothmax}. \\
Base constraint strength $\gamma$ & 1.0 & Base multiplier in ~\eqref{eq:app_gamma_definition}. \\
QP regularization $\lambda_{\mathrm{cbf}}$ & 0.01 & Joint-space regularization. \\
Task position weight & 1.0 & Position rows in $J_{\mathcal E}^\top WJ_{\mathcal E}$. \\
Task rotation weight & 0.1 & Rotation rows in $J_{\mathcal E}^\top WJ_{\mathcal E}$. \\
CBF update clip $\Delta q_{\max}^{\mathrm{cbf}}$ & 0.1 & Clipping applied to the guidance update. \\
\bottomrule
\end{tabular}
\end{table}

\subsection{Runtime Breakdown}
\label{app:runtime_breakdown}

\begin{figure}[h]
\centering
\begin{minipage}[c]{0.64\linewidth}
\centering
\scriptsize
\setlength{\tabcolsep}{2.2pt}
\renewcommand{\arraystretch}{1.08}
\begin{tabular}{p{0.28\linewidth}p{0.41\linewidth}cc}
\toprule
\textbf{Component} & \textbf{Included operations} & \textbf{ms} & \textbf{\%} \\
\midrule
\multicolumn{4}{l}{\textit{Base Cartesian policy components}} \\
Diffusion U-Net & U-Net denoising, 16 steps & 41.2 & 39.9 \\
Observation Encoder & Image/state observation encoding & 5.6 & 5.4 \\
\textbf{Base policy subtotal} & Measured base-policy components & \textbf{46.8} & \textbf{45.4} \\
\midrule
\multicolumn{4}{l}{\textit{Additional \textsc{EmbodiSteer} components}} \\
Kinematic Mapping & FK, Jacobian, DLS, twist & 13.5 & 13.1 \\
CBF Guidance & CBF autograd, QP, linearization setup & 11.5 & 11.2 \\
Collision SDF Query & Sphere FK, SDF query, world setup & 2.8 & 2.7 \\
Pose / Norm. / Scheduler & Pose conversion, normalizer, scheduler step & 11.1 & 10.8 \\
I/O and Action Formatting & Obs. preprocessing, joint tensor, action conversion & 1.0 & 1.0 \\
Other Runtime Overhead & Uninstrumented loop/PyTorch overhead & 16.3 & 15.8 \\
\textbf{Added subtotal} & Joint-space lifting and guidance overhead & \textbf{56.2} & \textbf{54.6} \\
\midrule
\textbf{Full total} & Guided joint-space inference & \textbf{103.0} & \textbf{100.0} \\
\bottomrule
\end{tabular}
\end{minipage}
\hfill
\begin{minipage}[c]{0.32\linewidth}
\centering
\includegraphics[width=\linewidth]{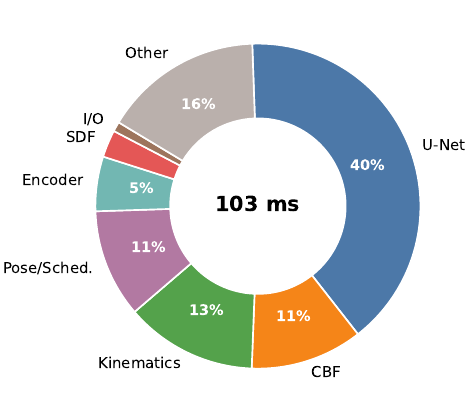}
\end{minipage}
\caption{Runtime breakdown for one guided joint-space inference call on an RTX
4070 Ti SUPER.}
\label{fig:runtime_breakdown}
\end{figure}

Fig.~\ref{fig:runtime_breakdown} reports the measured runtime breakdown of one
guided joint-space inference call on an RTX 4070 Ti SUPER, including all 16
reverse-diffusion steps. The full \textsc{EmbodiSteer} inference time is
103 ms, corresponding to approximately 9.7 Hz deployment. The base Cartesian policy already
requires observation encoding and U-Net denoising, which account for
46.8 ms. The remaining measured time is introduced by joint-space lifting,
whole-body SDF queries, CBF-QP guidance, and associated runtime overhead.

\subsection{Baseline Implementation Details}
\label{app:baselines}

Table~\ref{tab:baseline_details} summarizes how each baseline is implemented.
All baselines use the same trained Cartesian policy checkpoint for a given task.

\begin{table}[h]
\centering
\caption{Baseline implementation details.}
\label{tab:baseline_details}
\small
\setlength{\tabcolsep}{4pt}
\renewcommand{\arraystretch}{1.12}
\begin{tabular}{p{0.15\linewidth}p{0.75\linewidth}}
\toprule
\textbf{Method} & \textbf{Implementation} \\
\midrule
\textit{EE} &
Runs the Cartesian diffusion policy in its original action space and executes
the predicted end-effector actions without collision-aware guidance. \\
\textit{Joint} &
Uses the same joint-space denoising procedure as \textsc{EmbodiSteer}, including
FK queries to the frozen Cartesian denoiser and damped Jacobian updates, but
does not apply collision guidance. \\
\textit{EE w/ Sampling} &
Keeps Cartesian denoising unchanged, samples multiple candidate action chunks,
realizes them through the target robot, and selects the candidate with the
largest clearance. Number of candidates: 16. \\
\textit{EE w/ CBF} &
Keeps Cartesian denoising unchanged, solves IK for the generated action chunk,
and applies a one-shot CBF-QP correction after generation. \\
\textit{Joint w/ CG} &
Adds collision-cost gradient guidance during joint-space denoising. Unlike
\textsc{EmbodiSteer}, the update is not explicitly regularized by the
task-space QP objective, so whole-body collision gradients can perturb the
end-effector trajectory. \\
\bottomrule
\end{tabular}
\end{table}

\section{Simulation Results}
\label{app:simulation}

\subsection{Task Definitions and Evaluation Protocol}
\label{app:sim_tasks}

The simulation benchmark contains three manipulation tasks:
\textsc{PlaceToast}, \textsc{TurnOnFaucet}, and \textsc{MakeCoffee}. Cartesian
policies are trained from obstacle-free floating-gripper demonstrations
generated by task-specific motion planning, and are evaluated zero-shot on robot
embodiments with different kinematics and arm geometries. In obstacle-present
settings, obstacles are introduced only at test time. Success rate (TSR), maximum reward per episode (RWD), and collision occurrence rate (COR) are reported as in the
main text.

As illustrated in Fig.~\ref{fig:sim_exp_setup}, the benchmark uses the same
three tasks and obstacle layouts across 9 robot embodiments. We therefore avoid
duplicating the task and robot lists here; the appendix focuses on the complete
per-robot quantitative results and guidance ablations.

\subsection{Full Quantitative Results}
\label{app:sim_full_results}

Table~\ref{tab:sim_full_results} reports the per-robot simulation results
behind Table~\ref{tab:main_results}. For \textsc{EmbodiSteer}, the listed
guidance strengths match the main-paper setting for each task.

{\scriptsize
\setlength{\tabcolsep}{0.35pt}
\renewcommand{\arraystretch}{1.06}
\newcolumntype{Z}[1]{>{\centering\arraybackslash}m{#1}}
\begin{longtable}{@{}Z{2.55em}|Z{4.05em}|Z{4.75em}Z{3.90em}Z{3.05em}Z{3.45em}|Z{4.75em}Z{3.90em}Z{3.05em}Z{3.45em}|Z{4.75em}Z{3.90em}Z{3.05em}Z{3.45em}@{}}
\caption{Full per-robot simulation results. Obs. indicates whether obstacles are present during evaluation; TSR is success-once rate, RWD is maximum reward per episode, and COR is collision occurrence rate. COR is reported only for obstacle-present settings.}\label{tab:sim_full_results}\\
\toprule
\textbf{\makebox[0pt][c]{Obs.}} & \textbf{\makebox[0pt][c]{Method}} & \textbf{Robot} & \textbf{\makebox[0pt][c]{TSR$\uparrow$}} & \textbf{\makebox[0pt][c]{RWD$\uparrow$}} & \textbf{\makebox[0pt][c]{COR$\downarrow$}} & \textbf{Robot} & \textbf{\makebox[0pt][c]{TSR$\uparrow$}} & \textbf{\makebox[0pt][c]{RWD$\uparrow$}} & \textbf{\makebox[0pt][c]{COR$\downarrow$}} & \textbf{Robot} & \textbf{\makebox[0pt][c]{TSR$\uparrow$}} & \textbf{\makebox[0pt][c]{RWD$\uparrow$}} & \textbf{\makebox[0pt][c]{COR$\downarrow$}} \\
\midrule
\endfirsthead
\toprule
\textbf{\makebox[0pt][c]{Obs.}} & \textbf{\makebox[0pt][c]{Method}} & \textbf{Robot} & \textbf{\makebox[0pt][c]{TSR$\uparrow$}} & \textbf{\makebox[0pt][c]{RWD$\uparrow$}} & \textbf{\makebox[0pt][c]{COR$\downarrow$}} & \textbf{Robot} & \textbf{\makebox[0pt][c]{TSR$\uparrow$}} & \textbf{\makebox[0pt][c]{RWD$\uparrow$}} & \textbf{\makebox[0pt][c]{COR$\downarrow$}} & \textbf{Robot} & \textbf{\makebox[0pt][c]{TSR$\uparrow$}} & \textbf{\makebox[0pt][c]{RWD$\uparrow$}} & \textbf{\makebox[0pt][c]{COR$\downarrow$}} \\
\midrule
\endhead
\midrule
\multicolumn{14}{r}{\footnotesize Continued on next page} \\
\endfoot
\bottomrule
\endlastfoot
\multicolumn{14}{@{}l}{\textbf{\textsc{PlaceToast}}}\\\midrule
\multirow{8}{=}{\makecell{w/o\\Obs.}} & \multirow{4}{=}{\makecell{EE}} & UR5 & 100.0\% & 1.000 & -- & Panda & 100.0\% & 1.000 & -- & xArm6 & 100.0\% & 1.000 & -- \\*
  &  & xArm7 & 98.0\% & 0.989 & -- & iiwa7 & 98.0\% & 0.989 & -- & Gen3-6D & 81.0\% & 0.879 & -- \\*
  &  & Gen3-7D & 98.0\% & 0.989 & -- & Rizon4 & 99.0\% & 0.991 & -- & Sawyer & 97.0\% & 0.984 & -- \\*
\cmidrule(lr){11-14}
  &  &  &  &  &  &  &  &  &  & \textbf{Avg.} & \textbf{96.8\%} & \textbf{0.980} & -- \\
\cmidrule(lr){2-14}
  & \multirow{4}{=}{\makecell{Joint}} & UR5 & 100.0\% & 1.000 & -- & Panda & 100.0\% & 1.000 & -- & xArm6 & 98.0\% & 0.988 & -- \\*
  &  & xArm7 & 99.0\% & 0.995 & -- & iiwa7 & 98.0\% & 0.989 & -- & Gen3-6D & 73.0\% & 0.827 & -- \\*
  &  & Gen3-7D & 85.0\% & 0.907 & -- & Rizon4 & 98.0\% & 0.983 & -- & Sawyer & 99.0\% & 0.995 & -- \\*
\cmidrule(lr){11-14}
  &  &  &  &  &  &  &  &  &  & \textbf{Avg.} & \textbf{94.4\%} & \textbf{0.965} & -- \\
\cmidrule(lr){1-14}
 \multirow{20}{=}{\makecell{w/\\Obs.}} & \multirow{4}{=}{\makecell{EE}} & UR5 & 39.0\% & 0.560 & 50.0\% & Panda & 46.0\% & 0.585 & 54.0\% & xArm6 & 45.0\% & 0.676 & 52.0\% \\*
  &  & xArm7 & 48.0\% & 0.695 & 56.0\% & iiwa7 & 47.0\% & 0.689 & 49.0\% & Gen3-6D & 36.0\% & 0.611 & 51.0\% \\*
  &  & Gen3-7D & 43.0\% & 0.665 & 51.0\% & Rizon4 & 42.0\% & 0.552 & 51.0\% & Sawyer & 45.0\% & 0.676 & 59.0\% \\*
\cmidrule(lr){11-14}
  &  &  &  &  &  &  &  &  &  & \textbf{Avg.} & \textbf{43.4\%} & \textbf{0.634} & \textbf{52.6\%} \\
\cmidrule(lr){2-14}
  & \multirow{4}{=}{\makecell{EE w/\\Samp.}} & UR5 & 48.0\% & 0.626 & 62.0\% & Panda & 54.0\% & 0.651 & 46.0\% & xArm6 & 55.0\% & 0.736 & 49.0\% \\*
  &  & xArm7 & 52.0\% & 0.717 & 50.0\% & iiwa7 & 44.0\% & 0.663 & 40.0\% & Gen3-6D & 36.0\% & 0.613 & 47.0\% \\*
  &  & Gen3-7D & 45.0\% & 0.677 & 53.0\% & Rizon4 & 48.0\% & 0.601 & 62.0\% & Sawyer & 42.0\% & 0.661 & 53.0\% \\*
\cmidrule(lr){11-14}
  &  &  &  &  &  &  &  &  &  & \textbf{Avg.} & \textbf{47.1\%} & \textbf{0.661} & \textbf{51.3\%} \\
\cmidrule(lr){2-14}
  & \multirow{4}{=}{\makecell{EE w/\\CBF}} & UR5 & 76.0\% & 0.797 & 11.0\% & Panda & 64.0\% & 0.692 & 17.0\% & xArm6 & 75.0\% & 0.849 & 8.0\% \\*
  &  & xArm7 & 66.0\% & 0.796 & 14.0\% & iiwa7 & 62.0\% & 0.768 & 14.0\% & Gen3-6D & 62.0\% & 0.766 & 5.0\% \\*
  &  & Gen3-7D & 68.0\% & 0.809 & 13.0\% & Rizon4 & 50.0\% & 0.580 & 37.0\% & Sawyer & 69.0\% & 0.816 & 19.0\% \\*
\cmidrule(lr){11-14}
  &  &  &  &  &  &  &  &  &  & \textbf{Avg.} & \textbf{65.8\%} & \textbf{0.764} & \textbf{15.3\%} \\
\cmidrule(lr){2-14}
  & \multirow{4}{=}{\makecell{Joint\\w/ CG}} & UR5 & 35.0\% & 0.386 & 0.0\% & Panda & 76.0\% & 0.789 & 0.0\% & xArm6 & 62.0\% & 0.752 & 0.0\% \\*
  &  & xArm7 & 67.0\% & 0.786 & 0.0\% & iiwa7 & 44.0\% & 0.628 & 1.0\% & Gen3-6D & 54.0\% & 0.699 & 0.0\% \\*
  &  & Gen3-7D & 57.0\% & 0.721 & 0.0\% & Rizon4 & 62.0\% & 0.665 & 0.0\% & Sawyer & 62.0\% & 0.753 & 1.0\% \\*
\cmidrule(lr){11-14}
  &  &  &  &  &  &  &  &  &  & \textbf{Avg.} & \textbf{57.7\%} & \textbf{0.687} & \textbf{0.2\%} \\
\cmidrule(lr){2-14}
  & \multirow{4}{=}{\makecell{Ours}} & UR5 & 78.0\% & 0.807 & 0.0\% & Panda & 82.0\% & 0.842 & 3.0\% & xArm6 & 77.0\% & 0.859 & 0.0\% \\*
  &  & xArm7 & 68.0\% & 0.805 & 0.0\% & iiwa7 & 84.0\% & 0.905 & 9.0\% & Gen3-6D & 54.0\% & 0.712 & 0.0\% \\*
  &  & Gen3-7D & 76.0\% & 0.853 & 1.0\% & Rizon4 & 82.0\% & 0.849 & 16.0\% & Sawyer & 72.0\% & 0.831 & 12.0\% \\*
\cmidrule(lr){11-14}
  &  &  &  &  &  &  &  &  &  & \textbf{Avg.} & \textbf{74.8\%} & \textbf{0.829} & \textbf{4.6\%} \\
\midrule
\multicolumn{14}{@{}l}{\textbf{\textsc{TurnOnFaucet}}}\\\midrule
\multirow{8}{=}{\makecell{w/o\\Obs.}} & \multirow{4}{=}{\makecell{EE}} & Panda & 81.0\% & 0.943 & -- & xArm6 & 99.0\% & 0.998 & -- & xArm7 & 78.0\% & 0.937 & -- \\*
  &  & UR5 & 90.0\% & 0.918 & -- & iiwa7 & 87.0\% & 0.964 & -- & Gen3-6D & 96.0\% & 0.988 & -- \\*
  &  & Gen3-7D & 53.0\% & 0.880 & -- & Rizon4 & 73.0\% & 0.822 & -- & Sawyer & 95.0\% & 0.987 & -- \\*
\cmidrule(lr){11-14}
  &  &  &  &  &  &  &  &  &  & \textbf{Avg.} & \textbf{83.6\%} & \textbf{0.937} & -- \\
\cmidrule(lr){2-14}
  & \multirow{4}{=}{\makecell{Joint}} & UR5 & 88.0\% & 0.912 & -- & Panda & 90.0\% & 0.974 & -- & xArm6 & 100.0\% & 1.000 & -- \\*
  &  & xArm7 & 90.0\% & 0.971 & -- & iiwa7 & 91.0\% & 0.975 & -- & Gen3-6D & 95.0\% & 0.988 & -- \\*
  &  & Gen3-7D & 63.0\% & 0.909 & -- & Rizon4 & 87.0\% & 0.952 & -- & Sawyer & 87.0\% & 0.936 & -- \\*
\cmidrule(lr){11-14}
  &  &  &  &  &  &  &  &  &  & \textbf{Avg.} & \textbf{87.9\%} & \textbf{0.957} & -- \\
\cmidrule(lr){1-14}
 \multirow{20}{=}{\makecell{w/\\Obs.}} & \multirow{4}{=}{\makecell{EE}} & UR5 & 11.0\% & 0.417 & 86.0\% & Panda & 23.0\% & 0.471 & 94.0\% & xArm6 & 26.0\% & 0.791 & 89.0\% \\*
  &  & xArm7 & 41.0\% & 0.823 & 61.0\% & iiwa7 & 66.0\% & 0.915 & 28.0\% & Gen3-6D & 73.0\% & 0.931 & 41.0\% \\*
  &  & Gen3-7D & 42.0\% & 0.860 & 40.0\% & Rizon4 & 12.0\% & 0.657 & 88.0\% & Sawyer & 80.0\% & 0.904 & 29.0\% \\*
\cmidrule(lr){11-14}
  &  &  &  &  &  &  &  &  &  & \textbf{Avg.} & \textbf{41.6\%} & \textbf{0.752} & \textbf{61.8\%} \\
\cmidrule(lr){2-14}
  & \multirow{4}{=}{\makecell{EE w/\\Samp.}} & UR5 & 6.0\% & 0.453 & 86.0\% & Panda & 35.0\% & 0.554 & 87.0\% & xArm6 & 31.0\% & 0.811 & 84.0\% \\*
  &  & xArm7 & 32.0\% & 0.789 & 47.0\% & iiwa7 & 69.0\% & 0.911 & 18.0\% & Gen3-6D & 74.0\% & 0.935 & 28.0\% \\*
  &  & Gen3-7D & 50.0\% & 0.874 & 21.0\% & Rizon4 & 13.0\% & 0.665 & 90.0\% & Sawyer & 85.0\% & 0.911 & 18.0\% \\*
\cmidrule(lr){11-14}
  &  &  &  &  &  &  &  &  &  & \textbf{Avg.} & \textbf{43.9\%} & \textbf{0.767} & \textbf{53.2\%} \\
\cmidrule(lr){2-14}
  & \multirow{4}{=}{\makecell{EE w/\\CBF}} & UR5 & 33.0\% & 0.506 & 29.0\% & Panda & 48.0\% & 0.661 & 46.0\% & xArm6 & 60.0\% & 0.857 & 36.0\% \\*
  &  & xArm7 & 55.0\% & 0.833 & 26.0\% & iiwa7 & 78.0\% & 0.945 & 9.0\% & Gen3-6D & 87.0\% & 0.942 & 12.0\% \\*
  &  & Gen3-7D & 48.0\% & 0.851 & 1.0\% & Rizon4 & 8.0\% & 0.632 & 97.0\% & Sawyer & 90.0\% & 0.949 & 13.0\% \\*
\cmidrule(lr){11-14}
  &  &  &  &  &  &  &  &  &  & \textbf{Avg.} & \textbf{56.3\%} & \textbf{0.797} & \textbf{29.9\%} \\
\cmidrule(lr){2-14}
  & \multirow{4}{=}{\makecell{Joint\\w/ CG}} & UR5 & 8.0\% & 0.448 & 88.0\% & Panda & 7.0\% & 0.339 & 82.0\% & xArm6 & 6.0\% & 0.432 & 75.0\% \\*
  &  & xArm7 & 0.0\% & 0.239 & 65.0\% & iiwa7 & 12.0\% & 0.501 & 42.0\% & Gen3-6D & 33.0\% & 0.667 & 30.0\% \\*
  &  & Gen3-7D & 11.0\% & 0.481 & 23.0\% & Rizon4 & 2.0\% & 0.173 & 70.0\% & Sawyer & 14.0\% & 0.350 & 61.0\% \\*
\cmidrule(lr){11-14}
  &  &  &  &  &  &  &  &  &  & \textbf{Avg.} & \textbf{10.3\%} & \textbf{0.403} & \textbf{59.6\%} \\
\cmidrule(lr){2-14}
  & \multirow{4}{=}{\makecell{Ours}} & UR5 & 32.0\% & 0.502 & 23.0\% & Panda & 57.0\% & 0.802 & 42.0\% & xArm6 & 75.0\% & 0.900 & 30.0\% \\*
  &  & xArm7 & 52.0\% & 0.842 & 28.0\% & iiwa7 & 76.0\% & 0.926 & 17.0\% & Gen3-6D & 90.0\% & 0.971 & 7.0\% \\*
  &  & Gen3-7D & 61.0\% & 0.897 & 2.0\% & Rizon4 & 15.0\% & 0.744 & 88.0\% & Sawyer & 87.0\% & 0.916 & 7.0\% \\*
\cmidrule(lr){11-14}
  &  &  &  &  &  &  &  &  &  & \textbf{Avg.} & \textbf{60.6\%} & \textbf{0.833} & \textbf{27.1\%} \\
\midrule
\multicolumn{14}{@{}l}{\textbf{\textsc{MakeCoffee}}}\\\midrule
\multirow{8}{=}{\makecell{w/o\\Obs.}} & \multirow{4}{=}{\makecell{EE}} & UR5 & 94.0\% & 0.952 & -- & Panda & 94.0\% & 0.951 & -- & xArm6 & 87.0\% & 0.917 & -- \\*
  &  & xArm7 & 90.0\% & 0.943 & -- & iiwa7 & 85.0\% & 0.913 & -- & Gen3-6D & 93.0\% & 0.960 & -- \\*
  &  & Gen3-7D & 94.0\% & 0.965 & -- & Rizon4 & 88.0\% & 0.904 & -- & Sawyer & 82.0\% & 0.881 & -- \\*
\cmidrule(lr){11-14}
  &  &  &  &  &  &  &  &  &  & \textbf{Avg.} & \textbf{89.7\%} & \textbf{0.932} & -- \\
\cmidrule(lr){2-14}
  & \multirow{4}{=}{\makecell{Joint}} & UR5 & 94.0\% & 0.951 & -- & Panda & 93.0\% & 0.943 & -- & xArm6 & 91.0\% & 0.937 & -- \\*
  &  & xArm7 & 90.0\% & 0.936 & -- & iiwa7 & 92.0\% & 0.959 & -- & Gen3-6D & 95.0\% & 0.975 & -- \\*
  &  & Gen3-7D & 88.0\% & 0.927 & -- & Rizon4 & 80.0\% & 0.833 & -- & Sawyer & 84.0\% & 0.900 & -- \\*
\cmidrule(lr){11-14}
  &  &  &  &  &  &  &  &  &  & \textbf{Avg.} & \textbf{89.7\%} & \textbf{0.929} & -- \\
\cmidrule(lr){1-14}
 \multirow{20}{=}{\makecell{w/\\Obs.}} & \multirow{4}{=}{\makecell{EE}} & UR5 & 8.0\% & 0.277 & 85.0\% & Panda & 24.0\% & 0.408 & 60.0\% & xArm6 & 27.0\% & 0.536 & 70.0\% \\*
  &  & xArm7 & 28.0\% & 0.540 & 69.0\% & iiwa7 & 26.0\% & 0.548 & 59.0\% & Gen3-6D & 25.0\% & 0.497 & 45.0\% \\*
  &  & Gen3-7D & 24.0\% & 0.513 & 47.0\% & Rizon4 & 24.0\% & 0.385 & 44.0\% & Sawyer & 14.0\% & 0.401 & 46.0\% \\*
\cmidrule(lr){11-14}
  &  &  &  &  &  &  &  &  &  & \textbf{Avg.} & \textbf{22.2\%} & \textbf{0.456} & \textbf{58.3\%} \\
\cmidrule(lr){2-14}
  & \multirow{4}{=}{\makecell{EE w/\\Samp.}} & UR5 & 8.0\% & 0.243 & 87.0\% & Panda & 24.0\% & 0.393 & 63.0\% & xArm6 & 24.0\% & 0.539 & 71.0\% \\*
  &  & xArm7 & 30.0\% & 0.577 & 64.0\% & iiwa7 & 39.0\% & 0.635 & 56.0\% & Gen3-6D & 20.0\% & 0.490 & 56.0\% \\*
  &  & Gen3-7D & 24.0\% & 0.523 & 52.0\% & Rizon4 & 23.0\% & 0.384 & 56.0\% & Sawyer & 25.0\% & 0.516 & 35.0\% \\*
\cmidrule(lr){11-14}
  &  &  &  &  &  &  &  &  &  & \textbf{Avg.} & \textbf{24.1\%} & \textbf{0.478} & \textbf{60.0\%} \\
\cmidrule(lr){2-14}
  & \multirow{4}{=}{\makecell{EE w/\\CBF}} & UR5 & 19.0\% & 0.314 & 8.0\% & Panda & 62.0\% & 0.683 & 11.0\% & xArm6 & 58.0\% & 0.699 & 14.0\% \\*
  &  & xArm7 & 48.0\% & 0.615 & 2.0\% & iiwa7 & 66.0\% & 0.762 & 4.0\% & Gen3-6D & 67.0\% & 0.769 & 6.0\% \\*
  &  & Gen3-7D & 43.0\% & 0.574 & 4.0\% & Rizon4 & 39.0\% & 0.494 & 21.0\% & Sawyer & 38.0\% & 0.543 & 7.0\% \\*
\cmidrule(lr){11-14}
  &  &  &  &  &  &  &  &  &  & \textbf{Avg.} & \textbf{48.9\%} & \textbf{0.606} & \textbf{8.6\%} \\
\cmidrule(lr){2-14}
  & \multirow{4}{=}{\makecell{Joint\\w/ CG}} & UR5 & 13.0\% & 0.287 & 79.0\% & Panda & 18.0\% & 0.333 & 51.0\% & xArm6 & 29.0\% & 0.528 & 56.0\% \\*
  &  & xArm7 & 31.0\% & 0.519 & 36.0\% & iiwa7 & 17.0\% & 0.385 & 28.0\% & Gen3-6D & 31.0\% & 0.541 & 39.0\% \\*
  &  & Gen3-7D & 33.0\% & 0.544 & 36.0\% & Rizon4 & 5.0\% & 0.183 & 13.0\% & Sawyer & 25.0\% & 0.493 & 27.0\% \\*
\cmidrule(lr){11-14}
  &  &  &  &  &  &  &  &  &  & \textbf{Avg.} & \textbf{22.4\%} & \textbf{0.424} & \textbf{40.6\%} \\
\cmidrule(lr){2-14}
  & \multirow{4}{=}{\makecell{Ours}} & UR5 & 28.0\% & 0.387 & 3.0\% & Panda & 75.0\% & 0.798 & 4.0\% & xArm6 & 68.0\% & 0.757 & 3.0\% \\*
  &  & xArm7 & 64.0\% & 0.756 & 2.0\% & iiwa7 & 73.0\% & 0.815 & 0.0\% & Gen3-6D & 69.0\% & 0.795 & 2.0\% \\*
  &  & Gen3-7D & 56.0\% & 0.662 & 1.0\% & Rizon4 & 38.0\% & 0.481 & 9.0\% & Sawyer & 44.0\% & 0.574 & 1.0\% \\*
\cmidrule(lr){11-14}
  &  &  &  &  &  &  &  &  &  & \textbf{Avg.} & \textbf{57.2\%} & \textbf{0.670} & \textbf{2.8\%} \\
\end{longtable}
}

\subsection{Repeated Evaluations}
\label{app:repeated_evaluations}
Tab.~\ref{tab:repeated_task_results} reports \textsc{EmbodiSteer}
evaluations with three different random seeds for each task, using the same
trained task policy.
Each row averages over the nine robot embodiments, using the task success,
maximum reward per episode, and collision occurrence metrics of the main
benchmark.\par

\begin{table}[ht]
\centering

\caption{Repeated \textsc{EmbodiSteer} evaluations with three different
random seeds for each task.}
\label{tab:repeated_task_results}
\small
\setlength{\tabcolsep}{8pt}
\begin{tabular}{llccc}
\toprule
\textbf{Task} & \textbf{Evaluation} & \textbf{TSR (\%) $\uparrow$} & \textbf{RWD $\uparrow$} & \textbf{COR (\%) $\downarrow$} \\
\midrule
\textsc{MakeCoffee} & Run 1 & 57.2 & .670 & 2.8 \\
 & Run 2 & 57.9 & .677 & 3.4 \\
 & Run 3 & 53.8 & .649 & 4.2 \\
\midrule
\textsc{TurnOnFaucet} & Run 1 & 60.6 & .833 & 27.1 \\
 & Run 2 & 62.0 & .843 & 23.9 \\
 & Run 3 & 57.8 & .809 & 29.3 \\
\midrule
\textsc{PlaceToast} & Run 1 & 74.8 & .829 & 4.6 \\
 & Run 2 & 73.1 & .818 & 5.4 \\
 & Run 3 & 72.6 & .810 & 4.7 \\
\bottomrule
\end{tabular}
\end{table}

The aggregate mean and standard deviation of these three-seed
evaluations are reported for \textsc{EmbodiSteer} in
Tab.~\ref{tab:repeated_seed_results}.\par

\subsection{Additional Ablations and Sensitivity Analyses}
\label{app:sim_ablations}

\begin{figure}[h]
\centering
\includegraphics[width=\linewidth]{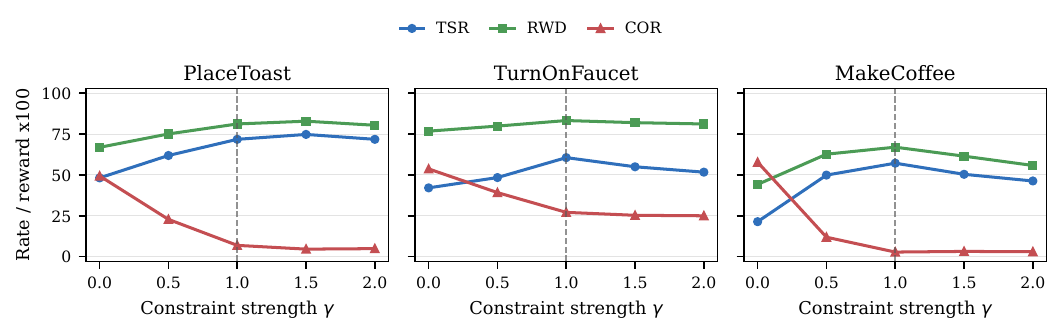}
\caption{CBF-QP guidance-strength sensitivity analysis. The default
$\gamma=1.0$ follows the linearized QP derivation and provides a robust
task--collision tradeoff across the three tasks.}
\label{fig:cbf_guidance_ablation}
\end{figure}

We first analyze the main design choices and hyperparameters of
\textsc{EmbodiSteer}. The constraint strength $\gamma$ controls the strength of the
CBF-QP constraint relative to the diffusion update: too small a value may leave
collisions unresolved, while too large a value can over-steer the sample and
hurt task execution. Following the linearized QP derivation, we use
$\gamma=1.0$ as the default scale and sweep around this value.
Fig.~\ref{fig:cbf_guidance_ablation} therefore visualizes a guidance-strength
sensitivity analysis across tasks.

The regularization weight $\lambda_{\mathrm{cbf}}$ controls how much the QP
penalizes joint-space deviation from the denoised sample. This term is included
to prevent collision avoidance from being absorbed by unnecessarily large joint
motions. Table~\ref{tab:cbf_lambda_ablation} shows that performance is stable
over a broad range, with the default $\lambda_{\mathrm{cbf}}=0.01$ giving a
balanced setting.

\begin{table}[h]
\centering
\caption{CBF-QP regularization sensitivity analysis averaged over the 9 robot embodiments used in the paper. The $\lambda_{\mathrm{cbf}}=0.01$ row corresponds to the default setting.}
\label{tab:cbf_lambda_ablation}
\small
\setlength{\tabcolsep}{6pt}
\renewcommand{\arraystretch}{1.08}
\begin{tabular}{l c c c c}
\toprule
\textbf{Task} & $\boldsymbol{\lambda_{\mathrm{cbf}}}$ & \textbf{TSR $\uparrow$} & \textbf{RWD $\uparrow$} & \textbf{COR $\downarrow$} \\
\midrule
\multirow{8}{*}{\textsc{PlaceToast}}
& 0.000 & 70.8\% & 0.800 & 5.9\% \\
& 0.001 & 70.0\% & 0.797 & 6.1\% \\
& 0.005 & \textbf{72.4\%} & 0.809 & \textbf{5.0\%} \\
& 0.010 & 71.8\% & \textbf{0.812} & 6.9\% \\
& 0.050 & 70.9\% & 0.797 & 5.3\% \\
& 0.100 & 70.1\% & 0.797 & 5.7\% \\
& 0.500 & 69.9\% & 0.797 & 5.9\% \\
& 1.000 & 70.9\% & 0.803 & 7.1\% \\
\midrule
\multirow{8}{*}{\textsc{TurnOnFaucet}}
& 0.000 & 57.2\% & 0.829 & 26.3\% \\
& 0.001 & 58.9\% & \textbf{0.842} & 25.7\% \\
& 0.005 & 58.6\% & 0.833 & 24.9\% \\
& 0.010 & \textbf{60.6\%} & 0.833 & 27.1\% \\
& 0.050 & 57.0\% & 0.817 & 26.6\% \\
& 0.100 & 59.6\% & 0.831 & 25.9\% \\
& 0.500 & 59.7\% & 0.838 & 25.7\% \\
& 1.000 & 60.2\% & 0.835 & \textbf{24.6\%} \\
\midrule
\multirow{8}{*}{\textsc{MakeCoffee}}
& 0.000 & 46.3\% & 0.603 & 3.6\% \\
& 0.001 & 42.4\% & 0.574 & 3.2\% \\
& 0.005 & 45.3\% & 0.591 & 2.6\% \\
& 0.010 & \textbf{57.2\%} & \textbf{0.670} & 2.8\% \\
& 0.050 & 47.2\% & 0.606 & \textbf{2.0\%} \\
& 0.100 & 42.7\% & 0.570 & 3.4\% \\
& 0.500 & 44.7\% & 0.590 & 3.6\% \\
& 1.000 & 46.4\% & 0.603 & 3.6\% \\
\bottomrule
\end{tabular}
\end{table}

We also ablate the guidance schedule by comparing the scheduled strength in
~\eqref{eq:app_gamma_definition} with a constant scale
$\gamma_t=\gamma=1.0$ throughout denoising. The schedule is intended to avoid
strong corrections in early noisy denoising steps, where samples are still far
from the policy distribution. Table~\ref{tab:guidance_schedule_ablation} shows
that scheduling helps preserve task performance while maintaining similar
collision rates.

\begin{table}[h]
\centering
\caption{Guidance schedule ablation averaged over the 9 robot embodiments used in the paper. Scheduled guidance uses Eq.~\eqref{eq:app_gamma_definition}; constant guidance keeps $\gamma_t=\gamma=1.0$ for all denoising steps.}
\label{tab:guidance_schedule_ablation}
\small
\setlength{\tabcolsep}{5pt}
\renewcommand{\arraystretch}{1.08}
\begin{tabular}{l c c c c}
\toprule
\textbf{Task} & \textbf{Guidance scale} & \textbf{TSR $\uparrow$} & \textbf{RWD $\uparrow$} & \textbf{COR $\downarrow$} \\
\midrule
\multirow{2}{*}{\textsc{PlaceToast}}
& Scheduled $\gamma_t$ & \textbf{71.8\%} & \textbf{0.812} & 6.9\% \\
& Constant $\gamma_t$ & 70.2\% & 0.796 & \textbf{5.6\%} \\
\midrule
\multirow{2}{*}{\textsc{TurnOnFaucet}}
& Scheduled $\gamma_t$ & \textbf{60.6\%} & \textbf{0.833} & 27.1\% \\
& Constant $\gamma_t$ & 57.8\% & 0.827 & \textbf{26.1\%} \\
\midrule
\multirow{2}{*}{\textsc{MakeCoffee}}
& Scheduled $\gamma_t$ & \textbf{57.2\%} & \textbf{0.670} & \textbf{2.8\%} \\
& Constant $\gamma_t$ & 46.2\% & 0.593 & \textbf{2.8\%} \\
\bottomrule
\end{tabular}
\end{table}

\begin{figure}[h]
\centering
\includegraphics[width=\linewidth]{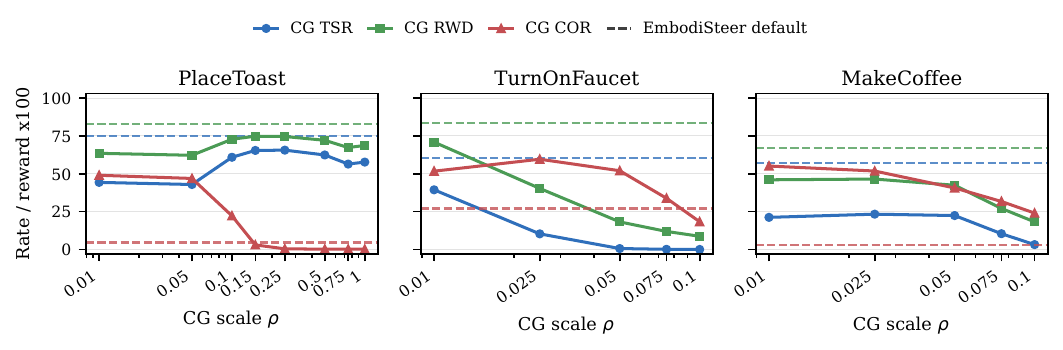}
\caption{Cost-gradient guidance-strength sensitivity analysis. Solid curves
show the CG baseline under different $\rho$, while dashed horizontal lines show
\textsc{EmbodiSteer} with its default setting. Tuning $\rho$ can reduce
collisions, but CG consistently gives a worse task--collision tradeoff.}
\label{fig:gd_guidance_ablation}
\end{figure}

Finally, we sweep the cost-gradient guidance scale $\rho$ for the gradient-based
baseline (\textit{Joint w/ CG}). This verifies that the baseline is not disadvantaged by a single
poorly chosen scale. As shown in Fig.~\ref{fig:gd_guidance_ablation}, tuning
$\rho$ can reduce collisions, but the unconstrained gradient update often
degrades task performance, especially in tasks requiring precise end-effector
motion.

\subsection{Obstacle-Present Training and Obstacle-Estimation Noise}
\label{app:obstacle_experiments}

\textbf{Obstacle-Present Training.}
The main benchmark trains Cartesian policies on obstacle-free demonstrations
and introduces surrounding obstacles at deployment. To test whether exposing
the Cartesian policy to obstacles during training can improve its subsequent
collision-aware task performance, we additionally train a policy on
obstacle-present demonstrations and evaluate it on \textsc{TurnOnFaucet}.
These demonstrations remain body-free: they can encode obstacle avoidance
for the end effector, but contain neither the target robot's joint
configurations nor its whole-body collision states. Thus, adding obstacles
to the demonstrations alone does not provide the embodiment-specific
information needed to learn whole-body avoidance for the deployed arm.
The obstacle-present training baseline increases TSR by 2.4 percentage points
over obstacle-free training,
while remaining 16.6 percentage points below \textsc{EmbodiSteer}. The
comparison indicates that obstacle-present training alone does not recover
the task performance achieved by joint-space guidance in this setting.\par

\textbf{Obstacle-Estimation Noise.}
Our guidance evaluates whole-body clearance against modeled cuboid obstacles.
To examine how errors in that geometric input affect execution, we perturb
the obstacle parameters supplied to guidance on \textsc{MakeCoffee}, while
keeping the evaluated scene unchanged. Translation and cuboid extent use
standard deviations of 5 or 10 mm; rotation uses a standard deviation of
$3^\circ$ in both noise settings. Tab.~\ref{tab:obstacle_estimation_noise}
shows that task success decreases and collision occurrence increases with
larger estimation noise.\par

\begin{table}[h]
\centering

\caption{Obstacle-estimation noise on \textsc{MakeCoffee}. Translation and
cuboid extent share the listed standard deviation; rotation standard deviation
is $3^\circ$ for both perturbed settings.}
\label{tab:obstacle_estimation_noise}
\small
\setlength{\tabcolsep}{12pt}
\begin{tabular}{lcc}
\toprule
\textbf{Translation / extent std} & \textbf{TSR (\%) $\uparrow$} & \textbf{COR (\%) $\downarrow$} \\
\midrule
No perturbation & 57.2 & 2.8 \\
5 mm & 50.4 & 4.4 \\
10 mm & 44.6 & 6.8 \\
\bottomrule
\end{tabular}
\end{table}

\subsection{Demonstration Coverage Settings}
\label{app:demo_coverage_settings}
The policies used in the main experiments are trained on 200 diverse
obstacle-free demonstrations. To separate the effects of data quantity
and motion diversity, the \textsc{MakeCoffee} ablations in
Fig.~\ref{fig:demo_coverage} retrain the Cartesian policy under two sweeps.
The quantity sweep uses 20, 50, 100, and 200 demonstrations. The diversity
sweep holds the demonstration count at 200 and scales the waypoint
randomization used by the demonstration motion planner to 10\%, 33\%,
67\%, and 100\% of the original data-generation setting. Thus, the
diversity percentages measure waypoint randomization, rather than the
fraction of demonstrations retained. Each trained policy is evaluated
with \textit{EE}, \textit{EE w/ CBF}, \textit{Joint w/ CG}, and
\textsc{EmbodiSteer}. Fig.~\ref{fig:demo_coverage} shows increasing success
with both data quantity and waypoint randomization, supporting broad
motion coverage for guided policy deployment.\par

\subsection{Candidate Sampling Budget}
\label{app:sampling_budget}
The \textit{EE w/ Sampling} baseline generates multiple Cartesian action
chunks from the frozen policy, realizes them on the target robot, and selects
the candidate with the largest whole-body clearance. To examine whether a
larger candidate pool can improve this strategy, we vary $K$ from 4 to 32
on \textsc{TurnOnFaucet}; the main benchmark uses $K=16$.
Tab.~\ref{tab:sampling_budget} reports success averaged over the nine robot
embodiments. Increasing the budget yields modest gains: even $K=32$ reaches
45.9\% TSR, compared with 60.6\% for \textsc{EmbodiSteer} in
Tab.~\ref{tab:main_results}.
Inference latency grows approximately linearly with $K$, reaching
5.55 s per chunk at $K=16$ and 10.97 s at $K=32$
(Tab.~\ref{tab:sampling_budget}), making larger candidate pools costly
for closed-loop replanning.\par

\begin{table}[ht]
\centering

\caption{Candidate-budget sensitivity of \textit{EE w/ Sampling} on
\textsc{TurnOnFaucet}, averaged over nine robot embodiments, and inference
latency (mean $\pm$ standard deviation) measured separately on
\textsc{PlaceToast} with Panda and an RTX 4090.}
\label{tab:sampling_budget}
\small
\setlength{\tabcolsep}{8pt}
\begin{tabular}{lrrrr}
\toprule
\textbf{Candidates $K$} & 4 & 8 & 16 & 32 \\
\midrule
\textbf{TSR (\%)} & 43.1 & 44.4 & 43.9 & 45.9 \\
\midrule
\textbf{Latency (s)} & $1.4803\pm0.0255$ & $2.8291\pm0.0536$ & $5.5529\pm0.0834$ & $10.9728\pm0.1641$ \\
\bottomrule
\end{tabular}
\end{table}

\subsection{Action-Chunk Continuity}
\label{app:chunk_continuity}
The QP corrects horizon states independently. We examine whether these
corrections change joint-trajectory continuity using paired guided and
unguided chunks on Panda and UR5 across the three tasks. Each setting uses
10 random seeds with one episode per seed, giving 60 episodes and 2,540 paired chunks
in total. Each pair shares the observation, current joint state, and
diffusion random state; only the guided chunk is executed. These are
trajectory diagnostics, rather than separate guided/unguided rollout
performance comparisons. The gripper is excluded from all metrics.\par

For adjacent configurations, we measure
$s_i=\|q_i-q_{i-1}\|_2/\sqrt{D}$, with $q_0$ the current configuration and
$D$ the number of arm joints. We report the 95th percentile over the
executed prefixes, including the current-state-to-first-action transition.
The second-difference ratio compares the maximum second joint difference
of guided and unguided trajectories over the executed prefix.
QP activity is measured at the final denoising iteration over that prefix.
Tab.~\ref{tab:continuity_results} shows that the 95th-percentile step changes
by at most 1.6\% in five of six settings; \textsc{TurnOnFaucet}/Panda
increases by 12.7\%.\par

\begin{table}[ht]
\centering

\caption{Continuity over executed action prefixes. P95 step is measured in
radians; ``ratio'' is guided/unguided. $r_2$ is the maximum second-difference
ratio. QP activity is the percentage of active horizon-point constraints.}
\label{tab:continuity_results}
\small
\setlength{\tabcolsep}{3.5pt}
\begin{tabular}{llccccc}
\toprule
\textbf{Task} & \textbf{Robot} & \makecell{\textbf{P95 step}\\\textbf{unguided}} & \makecell{\textbf{P95 step}\\\textbf{guided}} & \textbf{Ratio} & $\boldsymbol{r_2}$ & \makecell{\textbf{QP active}\\\textbf{(\%)}} \\
\midrule
\textsc{MakeCoffee} & Panda & 0.0348 & 0.0351 & 1.011 & 1.000 & 10.5 \\
 & UR5 & 0.0216 & 0.0216 & 0.999 & 4.582 & 28.7 \\
\textsc{TurnOnFaucet} & Panda & 0.0392 & 0.0442 & 1.127 & 1.000 & 28.8 \\
 & UR5 & 0.0135 & 0.0135 & 1.003 & 1.292 & 54.5 \\
\textsc{PlaceToast} & Panda & 0.0244 & 0.0246 & 1.007 & 1.000 & 7.7 \\
 & UR5 & 0.0171 & 0.0174 & 1.016 & 2.693 & 7.2 \\
\bottomrule
\end{tabular}
\end{table}

Overall, these results show that collision guidance largely
preserves joint-step continuity across tasks and embodiments. Although the
maximum second-difference ratios reach 4.58 and 2.69 on UR5 for
\textsc{MakeCoffee} and \textsc{PlaceToast}, respectively, the P95 step
sizes in these settings change by at most 1.6\%, indicating that the bulk
of the joint-step distribution remains stable under guidance.\par

\subsection{Additional Qualitative Results}
\label{app:sim_qual}

\begin{figure}[h]
\centering
\includegraphics[width=\linewidth]{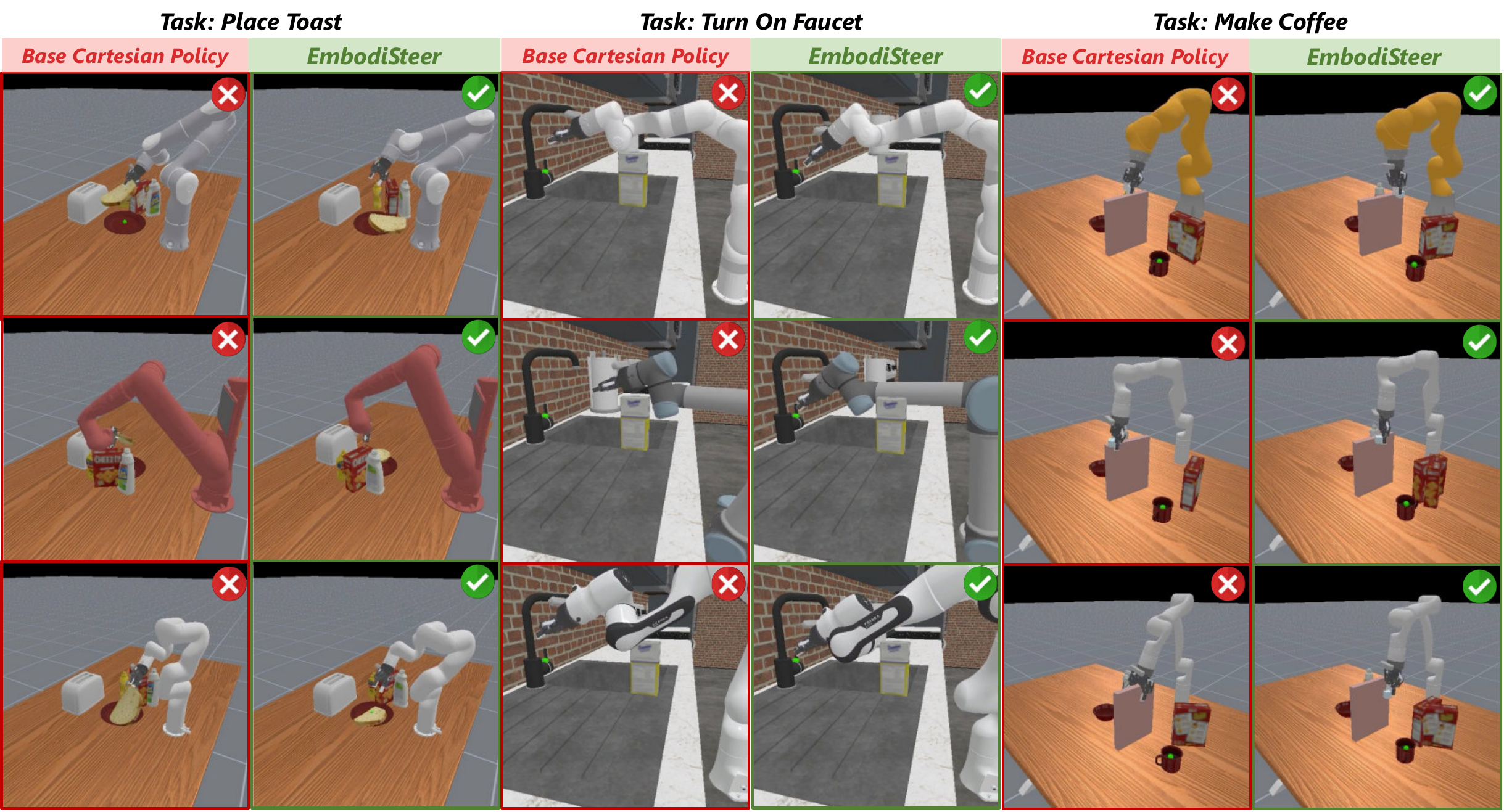}
\caption{Additional simulation qualitative results. The base Cartesian policy
can collide at the end-effector or along the arm body when obstacles are added
at test time, while \textsc{EmbodiSteer} generates collision-aware whole-body
motion and completes the task.}
\label{fig:app_sim_qual}
\end{figure}

Fig. ~\ref{fig:app_sim_qual} shows representative simulation rollouts across
the three benchmark tasks. The base Cartesian policy often produces
task-directed end-effector motion but cannot reason about out-of-distribution
obstacles, which leads to end-effector collisions near the task object. It also
ignores the target robot's arm geometry, leading to arm-body obstacle contact.
In contrast, \textsc{EmbodiSteer} preserves the same task intent while steering
both the end-effector and the full robot body around the obstacle.

\section{Real-World Results}
\label{app:real_world}

\subsection{Setup and Annotation Protocol}
\label{app:real_protocol}

\begin{figure}[h]
\centering
\includegraphics[width=0.8\linewidth]{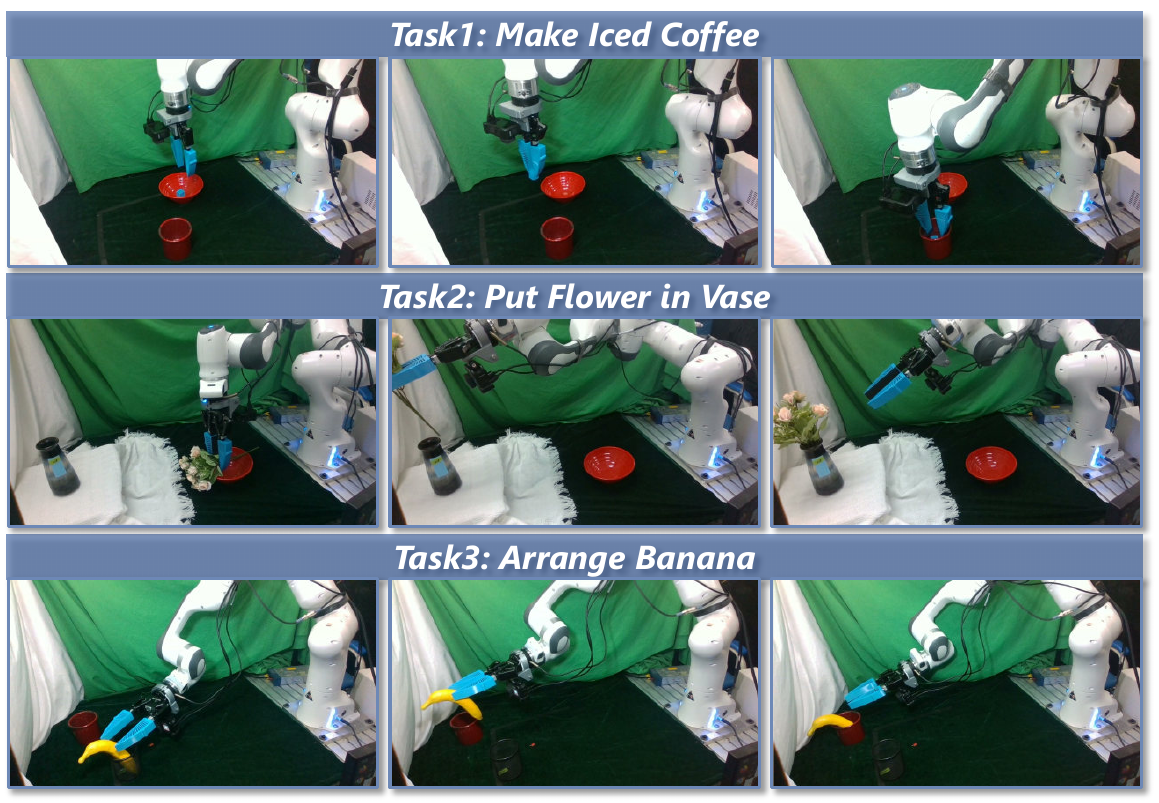}
\caption{Real-world task protocols. Each row shows representative stages of
one UMI-trained task: \textsc{MakeIcedCoffee}, \textsc{PutFlowerInVase}, and
\textsc{ArrangeBanana}.}
\label{fig:app_realworld_task}
\end{figure}

Each real-world task uses 200 UMI handheld-gripper demonstrations collected
without extra obstacles. Because these demonstrations are collected with a
handheld gripper rather than a particular robot arm, the training data are
embodiment-agnostic. For each task, a single Cartesian diffusion policy
checkpoint is trained and shared by UR5 and Franka Panda without robot-specific
finetuning. The three tasks are illustrated in
Fig.~\ref{fig:app_realworld_task}: \textsc{MakeIcedCoffee} requires grasping
blocks from a bowl, which represent ice cubes, and placing them into a cup;
\textsc{PutFlowerInVase} requires grasping the bouquet and inserting it into the
vase; and \textsc{ArrangeBanana} requires grasping a banana from a pen holder
and placing it into a cup. For each robot-task pair, we evaluate 10 no-obstacle
trials and 10 obstacle-present trials distributed across the obstacle types in
Fig.~\ref{fig:app_realworld_obstacle}. A trial is marked successful if the task
object reaches the intended final state without timeout, object drop, or human
intervention.

Task success and collision are annotated independently. For safety in the
real-world setup, obstacles are movable: after contact, they can be knocked over
or pushed aside, unlike the fixed obstacles in simulation that often stop the
robot motion. As a result, a rollout can still finish the task after a collision. COR
therefore measures whether any end-effector or arm-body obstacle contact occurs
during the rollout, separately from task success.

\subsection{Full Real-World Results}
\label{app:real_results}

Table~\ref{tab:real_world_full} reports the real-world results by robot and
task. Each task checkpoint is shared by both robots.

\begin{table}[h]
\centering
\caption{Real-world results by robot and task. TSR and COR are reported as
success/collision counts over 10 trials for each robot-task pair; the total row
aggregates over 60 trials.}
\label{tab:real_world_full}
\small
\setlength{\tabcolsep}{4pt}
\renewcommand{\arraystretch}{1.12}
\begin{tabular}{c c c c c c c}
\toprule
\multirow{2}{*}{\textbf{Robot}} & \multirow{2}{*}{\textbf{Task}} &
\multicolumn{1}{c}{\textbf{No Obstacle}} &
\multicolumn{2}{c}{\textbf{Obstacle, Base Policy}} &
\multicolumn{2}{c}{\textbf{Obstacle, \textsc{EmbodiSteer}}} \\
\cmidrule(lr){3-3}\cmidrule(lr){4-5}\cmidrule(lr){6-7}
& & \textbf{TSR $\uparrow$} & \textbf{TSR $\uparrow$} &
\textbf{COR $\downarrow$} & \textbf{TSR $\uparrow$} &
\textbf{COR $\downarrow$} \\
\midrule
\multirow{3}{*}{UR5} & \textsc{MakeIcedCoffee} & 9/10 & 4/10 & 10/10 & \textbf{8/10} & \textbf{1/10} \\
& \textsc{PutFlowerInVase} & 7/10 & 4/10 & 9/10 & \textbf{6/10} & \textbf{1/10} \\
& \textsc{ArrangeBanana} & 8/10 & 4/10 & 9/10 & \textbf{9/10} & \textbf{0/10} \\
\midrule
\multirow{3}{*}{Panda} & \textsc{MakeIcedCoffee} & 9/10 & 6/10 & 10/10 & \textbf{8/10} & \textbf{1/10} \\
& \textsc{PutFlowerInVase} & 6/10 & 1/10 & 10/10 & \textbf{6/10} & \textbf{0/10} \\
& \textsc{ArrangeBanana} & 10/10 & 6/10 & 10/10 & \textbf{10/10} & \textbf{1/10} \\
\midrule
Total & All tasks & 49/60 & 25/60 & 58/60 & \textbf{47/60} & \textbf{4/60} \\
\bottomrule
\end{tabular}
\end{table}

\subsection{Obstacle Layouts and Qualitative Results}
\label{app:real_qual}

\begin{figure}[h]
\centering
\includegraphics[width=\linewidth]{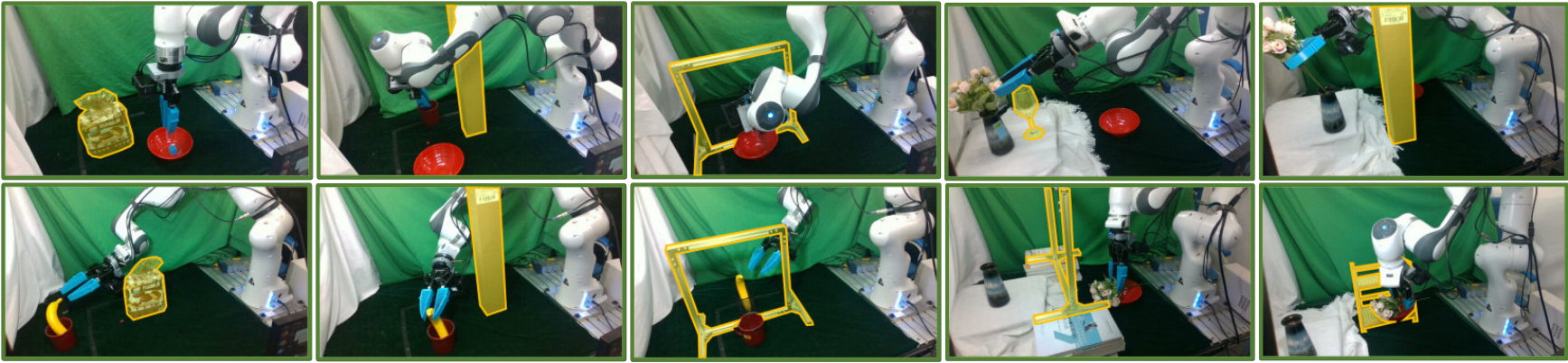}
\caption{Real-world obstacle layouts. The obstacles are highlighted with yellow masks. The obstacle type and pose are varied across episodes.}
\label{fig:app_realworld_obstacle}
\end{figure}

The real-world obstacle layouts are shown in
Fig.~\ref{fig:app_realworld_obstacle}. The obstacle type and pose are varied across episodes. These layouts require the robot to avoid collisions with both the arm links and the end effector. 

\begin{figure}[p]
\centering
\includegraphics[width=0.95\linewidth]{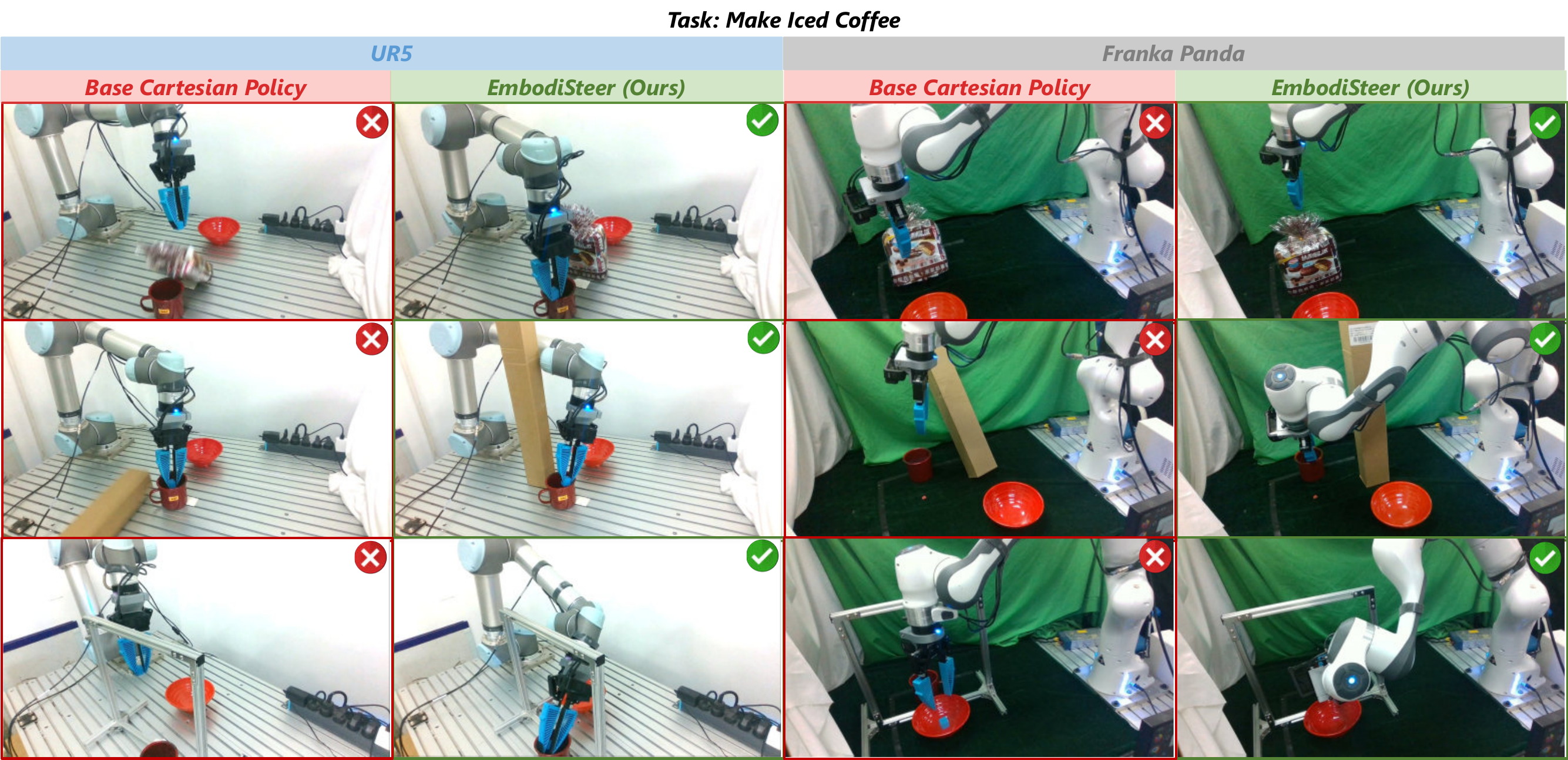}\\[-0.2em]
\includegraphics[width=0.95\linewidth]{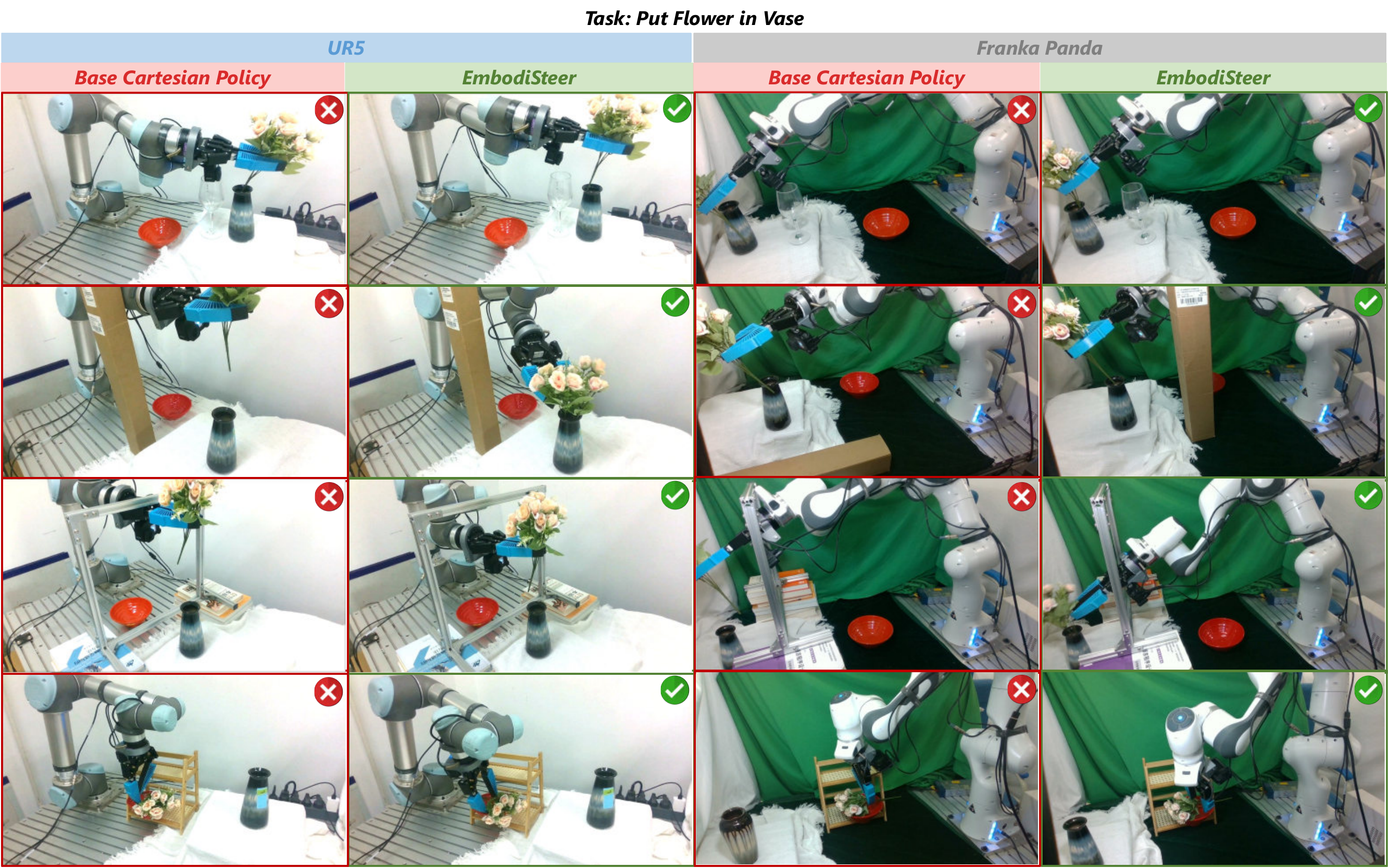}\\[-0.2em]
\includegraphics[width=0.95\linewidth]{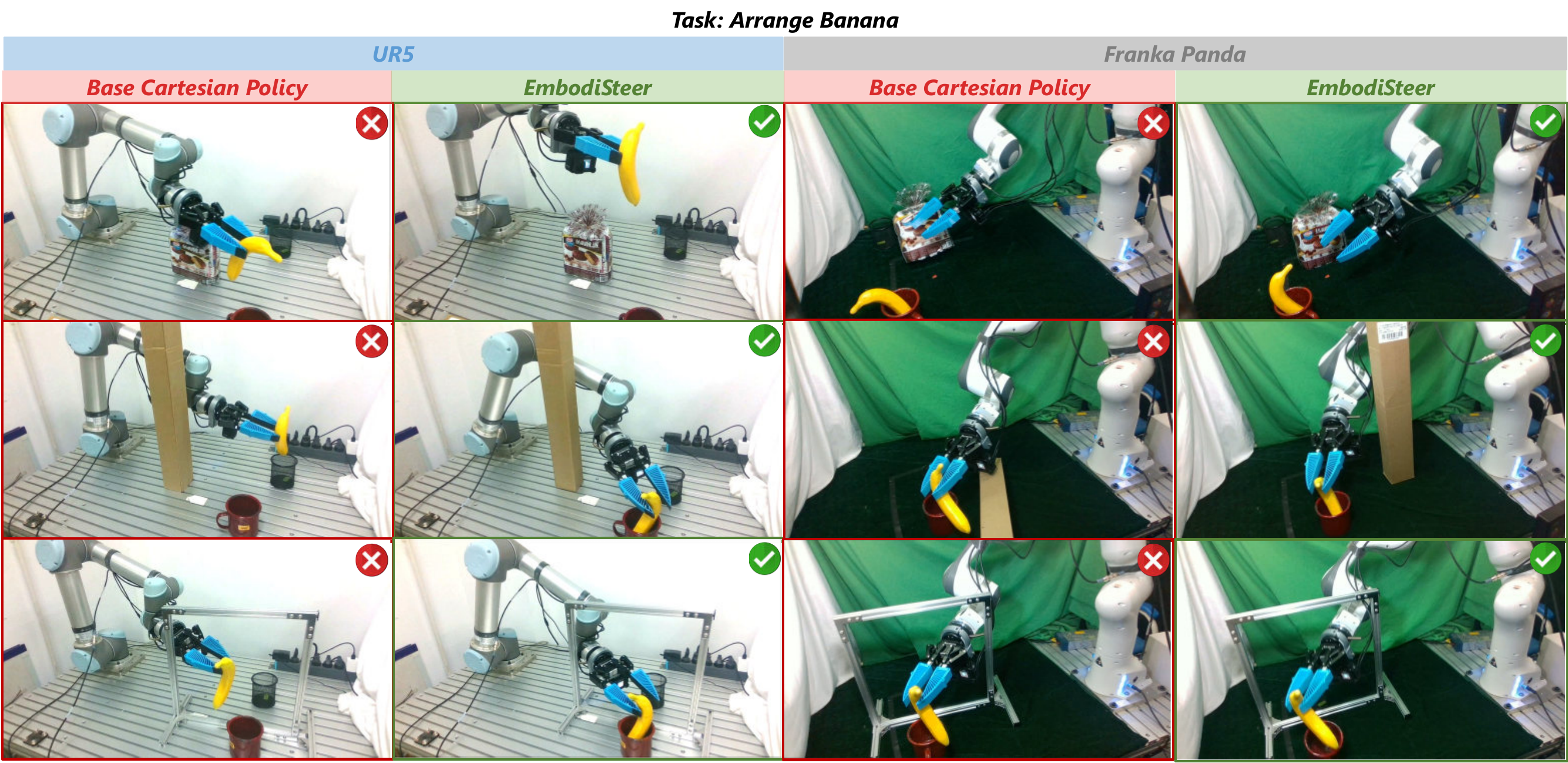}
\caption{Real-world qualitative results. From top to bottom:
\textsc{MakeIcedCoffee}, \textsc{PutFlowerInVase}, and
\textsc{ArrangeBanana}. Each row compares the base Cartesian policy and
\textsc{EmbodiSteer} on UR5 and Franka Panda under obstacle layouts that test
end-effector and arm-body avoidance.}
\label{fig:app_real_qual}
\end{figure}

Fig. ~\ref{fig:app_real_qual} shows representative real-world rollouts for all
three tasks. Each row compares the base Cartesian policy and
\textsc{EmbodiSteer} on both UR5 and Franka Panda. The qualitative results
mirror the quantitative trend in
Table~\ref{tab:real_world_full}: the base policy often reaches toward the task
object while contacting the obstacle, whereas \textsc{EmbodiSteer} modifies the
whole-body motion to avoid end-effector and arm-body obstacles.

\begin{figure}
\centering
\includegraphics[width=\linewidth]{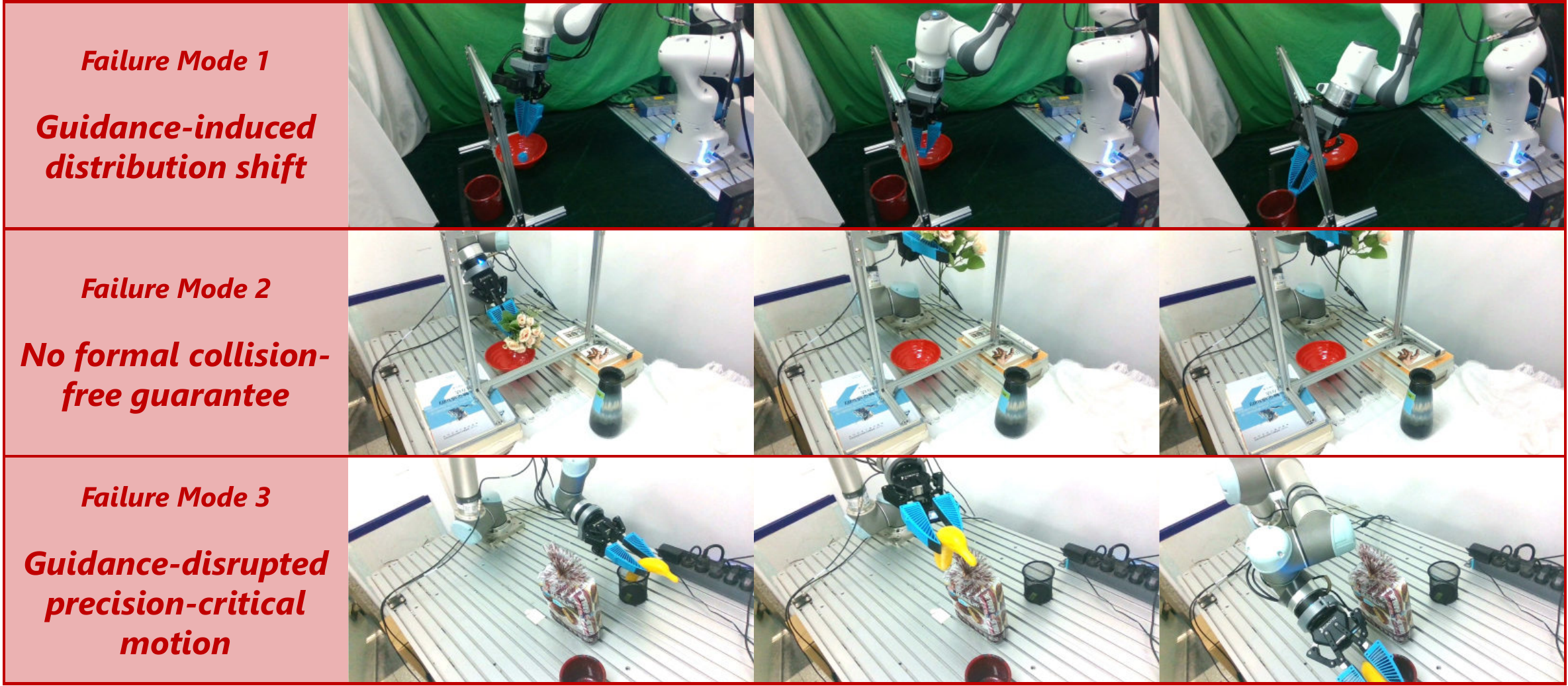}
\caption{Representative failure modes. Guidance may move the robot into
out-of-distribution states, fail to fully prevent collision under local
linearization and clipping, or disrupt precision-critical pick/place motions.}
\label{fig:app_failure_modes}
\end{figure}

\section{Failure Mode Analysis}
\label{app:failure_analysis}

Fig. ~\ref{fig:app_failure_modes} summarizes three representative failure modes
observed in our experiments. First, although guidance improves collision
avoidance on average, a large correction can steer the robot into joint states
that are rarely covered by the demonstrations. The frozen Cartesian policy may
then receive observations outside its training distribution and fail to recover
task progress. Second, the CBF-QP update is based on a local linearization and
clipped joint correction, so it provides a practical avoidance direction rather
than a formal collision-free certificate; collisions can still occur in highly
constrained layouts or when the required correction is too large. Third, some
task phases are more sensitive to small end-effector deviations than others. In
pick-and-place tasks, for example, guidance applied near grasping or placement
can perturb precision-critical motion and cause grasp misses, drops, or
misplacement, even when the resulting motion avoids obstacles.

\end{document}